%% file: guided_ig.tex
\documentclass[final]{cvpr}

\usepackage{times}
\usepackage{epsfig}
\usepackage{graphicx}
\usepackage{amsmath}
\usepackage{amssymb}
\usepackage{algorithm}
\usepackage{algpseudocode}
\usepackage[font={footnotesize}]{subcaption}
\usepackage{makecell}
\usepackage{booktabs}
\usepackage[font=small,belowskip=-4pt,aboveskip=3pt]{caption}

\usepackage[pagebackref=true,breaklinks=true,colorlinks,bookmarks=false]{hyperref}

\def\cvprPaperID{10338} %
\def\confYear{CVPR 2021}

\begin{document}

\title{Guided Integrated Gradients: an Adaptive Path Method for Removing Noise}

\author{Andrei Kapishnikov,
Subhashini Venugopalan,
Besim Avci,
Ben Wedin,
Michael Terry,
Tolga Bolukbasi\\
Google Research\\
{\tt\small \{kapishnikov, vsubhashini, besim, wedin, michaelterry, tolgab\}@google.com}
}

\maketitle

\begin{abstract}
Integrated Gradients (IG)~\cite{Sundararajan2017AxiomaticAF} is a commonly used feature attribution method for deep neural networks.
While IG has many desirable properties, the method often produces spurious/noisy pixel attributions in regions that are not related to the predicted class when applied to visual models. 
While this has been previously noted~\cite{sturmfels2020visualizing}, most existing solutions~\cite{Smilkov2017SmoothGradRN,miglani2020investigating} are aimed at addressing the symptoms by explicitly reducing the noise in the resulting attributions. In this work, we show that one of the causes of the problem is the accumulation of noise along the IG path. To minimize the effect of this source of noise, we propose adapting the attribution path itself - conditioning the path not just on the image but also on the model being explained. We introduce Adaptive Path Methods (APMs) as a generalization of path methods, and Guided IG as a specific instance of an APM. Empirically, Guided IG creates saliency maps better aligned with the model's prediction and the input image that is being explained. We show through qualitative and quantitative experiments that Guided IG outperforms other, related methods in nearly every experiment.

\end{abstract}

\section{Introduction}
\input{tex/introduction.tex}

\section{Related Work}
\input{tex/related_works.tex}

\section{High Gradient Impact on IG Attribution}
\input{tex/gradient_section}

\section{Adaptive Paths and Guided IG}

\input{tex/adaptive.tex}

\subsection{Desired Characteristics} %
\input{tex/guided.tex}

\section{Experiments and Results}

\input{tex/results}

\vspace{-5pt}
\section{Discussion}
\vspace{-5pt}

\input{tex/discussion}

\vspace{-5pt}
\section{Conclusion}
\vspace{-5pt}
\input{tex/conclusion}\\

\vspace{-4pt}
\noindent\textbf{Acknowledgements}
We thank Frederick Liu for his technical contributions to the saliency evaluation framework, experiments and feedback, and members of the Google Brain team for feedback on a draft of this work.

{\small
\bibliographystyle{ieee_fullname}
\bibliography{guided_ig}
}

\appendix
\clearpage
\input{tex/appendix}

\end{document}

%% file: tex/introduction.tex
As deep neural network computer vision models are integrated into critical applications such as healthcare and security, research on explaining these models has intensified. Feature attribution techniques 
strive to explain which inputs the model considers to be most important for a given prediction, making them useful tools in debugging models or understanding what they have likely learned. %
However, while a plethora of techniques have been developed~\cite{SVZ13,Sundararajan2017AxiomaticAF,kim2017interpretability,fong2017interpretable,ribeiro2016should}, there are still behaviors of these attribution techniques that remain to be understood. %
In this context, our work is focused on studying the source of noise in attributions produced by path-integral-based methods~\cite{sturmfels2020visualizing}.

Gradient-based feature attribution techniques~\cite{SVZ13,selvaraju2017grad} are of particular interest in our work.
The main idea behind these techniques is that the partial derivative of the output with respect to the input is considered as a measure of the sensitivity of the network for each input dimension. While early methods~\cite{SVZ13} use the gradients multiplied by the input as a feature attribution technique, more recent methods exploit gradients of the activation maps~\cite{selvaraju2017grad}, or integrate gradients over a path~\cite{Sundararajan2017AxiomaticAF}. %
This work studies Integrated Gradients (IG)~\cite{Sundararajan2017AxiomaticAF}, a commonly used method that is based on game-theoretic ideas in ~\cite{AumannShapley}. IG avoids the problem of diminishing influences of features due to gradient saturation, and has desirable theoretical properties.

\begin{figure}
    \centering
    \begin{subfigure}{0.31\columnwidth}
        \centering
        \includegraphics[width=\textwidth]{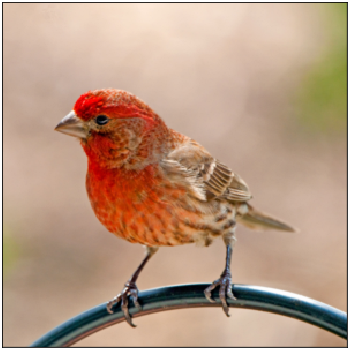}
        \caption{Input image}
        \label{fig:house_finch}
    \end{subfigure}
    \hfill
    \begin{subfigure}{0.31\columnwidth}
        \centering
        \includegraphics[width=\textwidth]{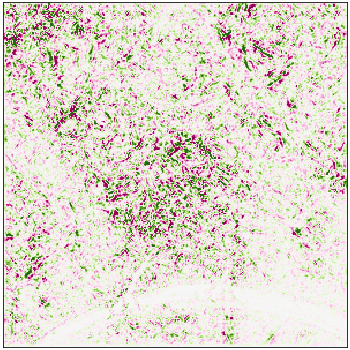}
        \caption{IG attributions}
        \label{fig:ig_attr}
    \end{subfigure}
    \hfill
    \begin{subfigure}{0.31\columnwidth}
        \centering
        \includegraphics[width=\textwidth]{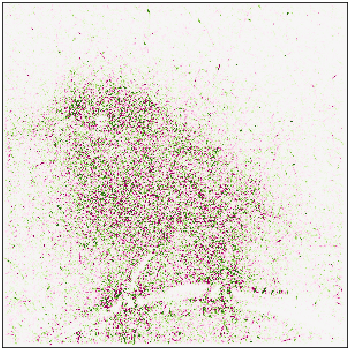}
        \caption{Guided IG}
        \label{fig:gig_attr}
    \end{subfigure}
    \caption{Comparing feature attribution for Integrated Gradients and Guided Integrated Gradients. Both (b) and (c) use a black baseline to explain the ``house finch'' prediction. While (b) has attributions on the bird, there is substantial noise in the attributions compared to (c). This work studies the source of noise, and presents Guided IG as a solution.}
    \label{fig:noise}
    \vspace{-0.3cm}
\end{figure}

\begin{figure*}
    \centering
    \begin{subfigure}{0.24\linewidth}
        \centering
        \includegraphics[width=\textwidth]{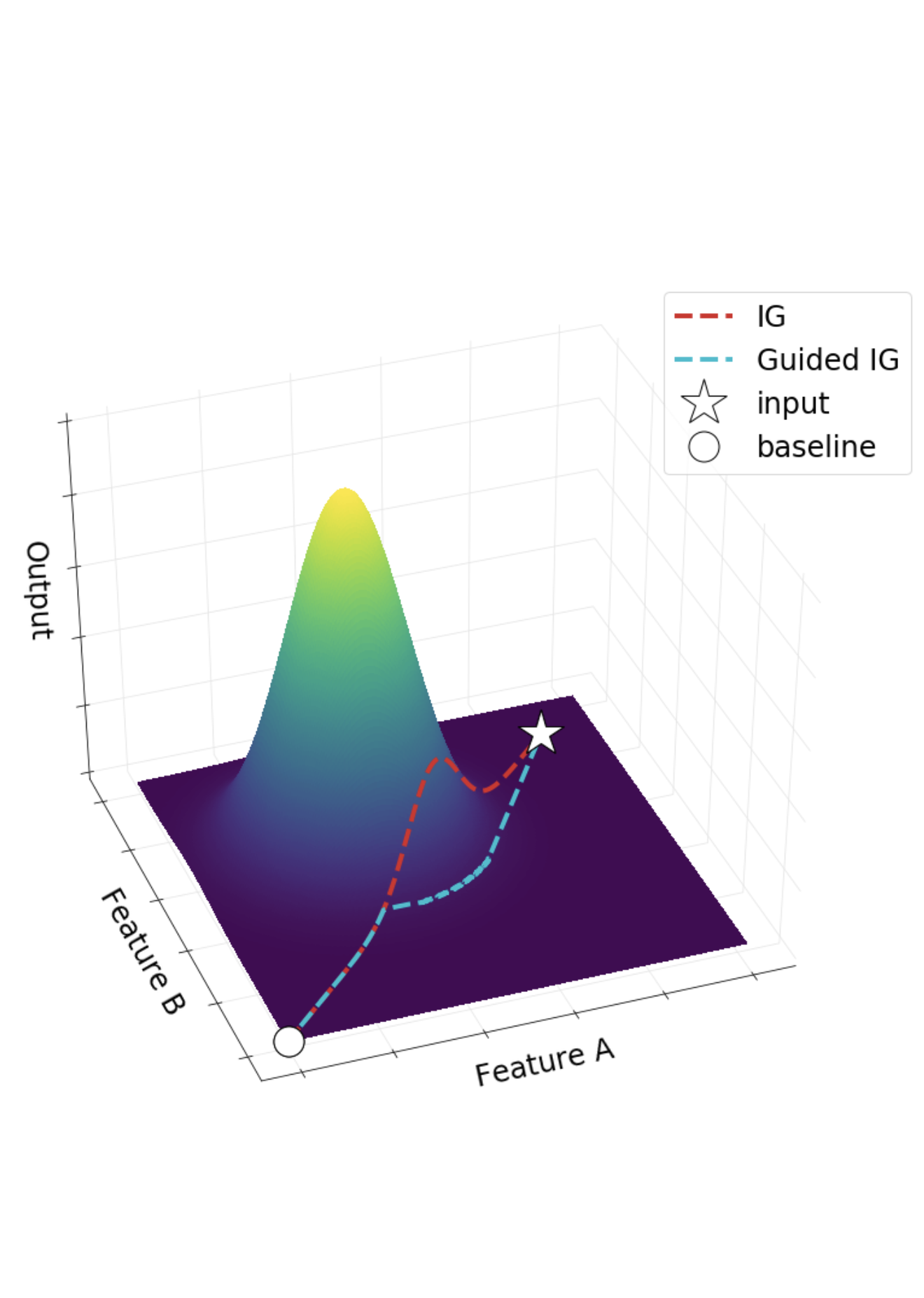}
        \caption{Schematic}
  \label{fig:ig-gig-paths}
    \end{subfigure}
    \hfill
    \begin{subfigure}{0.37\linewidth}
        \centering
        \includegraphics[width=\textwidth]{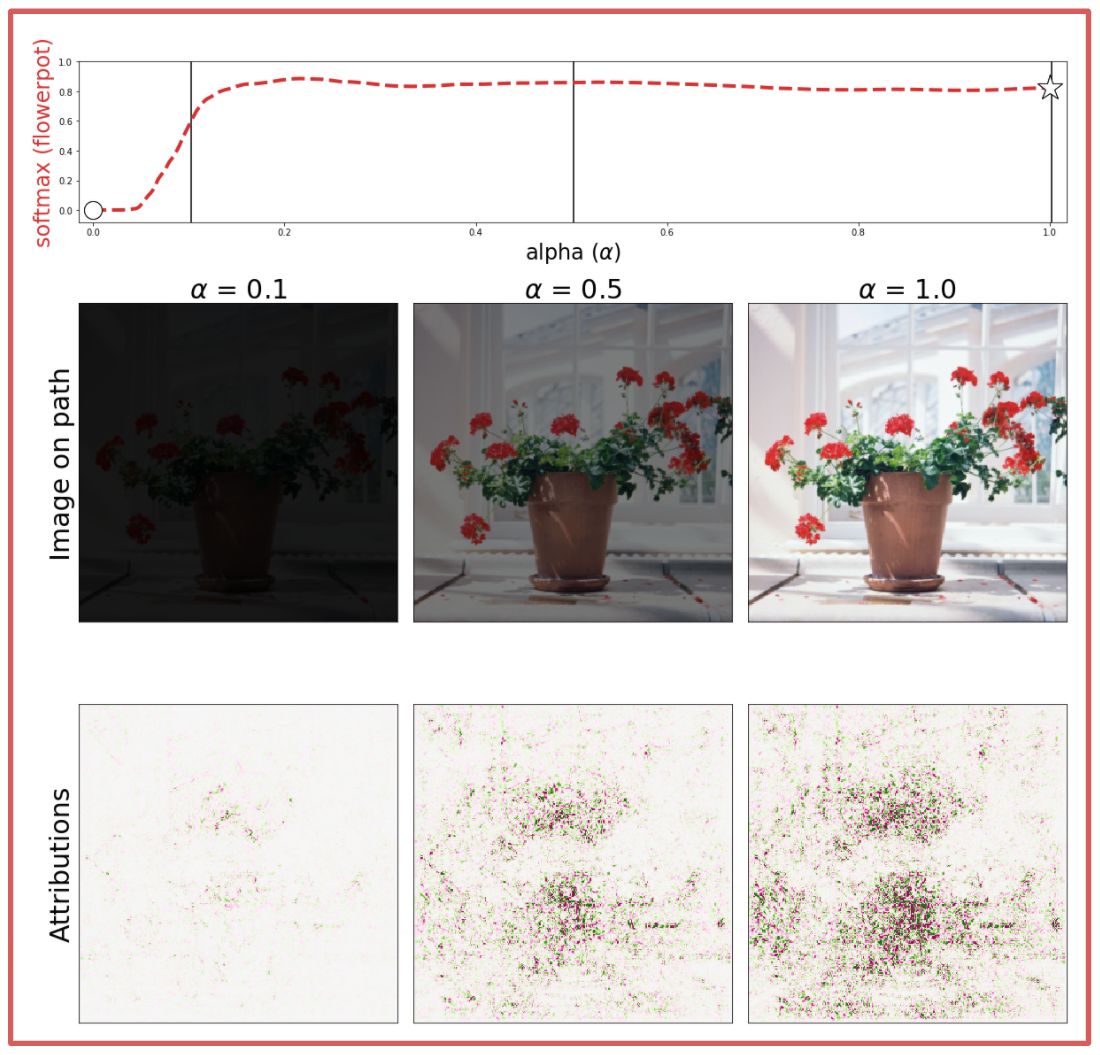}
        \caption{IG attributions}
        \label{fig:splash_ig_attr}
    \end{subfigure}
    \hfill
    \begin{subfigure}{0.37\linewidth}
        \centering
        \includegraphics[width=\textwidth]{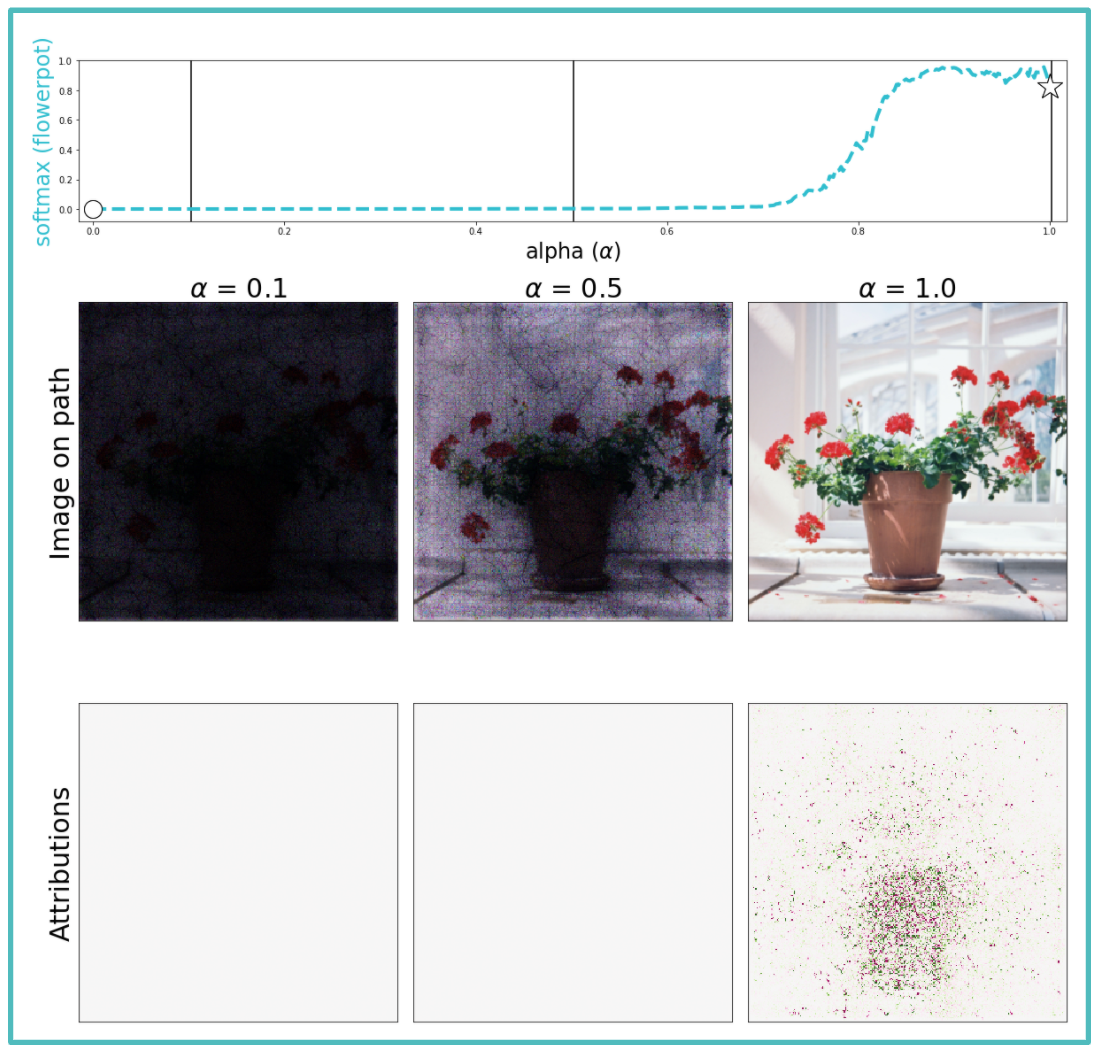}
        \caption{Guided IG attributions}
        \label{fig:splash_gig_attr}
    \end{subfigure}
  \caption{Comparing IG and Guided IG's paths and results. \textbf{(a)}: For IG, a straight line path from baseline to input is followed (red dotted line), regardless of changes in gradients. For Guided IG, the path is chosen by selecting features that have the smallest absolute value of corresponding partial derivatives (cyan dotted line). Guided IG's goal is to reduce the accumulation of gradients caused by nearby very high/very low prediction examples. \textbf{(b)} and \textbf{(c)}: Snapshots of attributions for the flower pot class for Integrated Gradients (center) and Guided IG (right) at alpha values of $0.1$, $0.5$, and $1.0$. The top rows show graphs of the softmax prediction for flower pot as a function of alpha. The second row shows the input image produced by each technique at the three different alpha values. Note that IG's straight line path affects all pixels equally (e.g., see $\alpha = 0.5$), while Guided IG reveals the least important features, first. The third row shows each technique's attributions for each of the three alpha values, with Guided IG showing less noise outside the area of the image occupied by the flower pot.}
  \label{fig:splash}
  \vspace{-0.3cm}
\end{figure*}

One commonly observed problem while calculating Integrated Gradients for vision models is the noise in pixel attribution (Figure \ref{fig:noise}) originating from gradient accumulation~\cite{Ghorbani2017,Smilkov2017SmoothGradRN,sturmfels2020visualizing, Xu_2020_CVPR} along the integration path. A few possible explanations for this noise have been put forth: (a) high curvature of the output manifold~\cite{Dombrowski2019ExplanationsCB}; (b) approximation of the integration with Riemann sum; and (c) choice of baselines~\cite{Xu_2020_CVPR, sturmfels2020visualizing}. 
Our experiments indicate that one source of attribution noise comes from regions of correlated, high-magnitude gradients on irrelevant pixels found along the straight line integration path. Our findings correlate with observations in ~\cite{Dombrowski2019ExplanationsCB}, which state that the model surface plays a large role in determining the magnitude of attribution values. 

Methods have been proposed to explicitly reduce the noise in attributions. SmoothGrad~\cite{Smilkov2017SmoothGradRN} averages attributions over multiple samples of the input, created by adding Gaussian noise to the original input. The aggregation improves the overall true signal in the attribution. %
XRAI~\cite{kapishnikov2019xrai} aggregates attributions within segments to reduce the outlier effect.  Sturmfels et al.  suggest the choice in baseline is a contributing factor, and propose different baselines as a potential solution~\cite{sturmfels2020visualizing}.  ~\cite{Xu_2020_CVPR} integrate over the frequency dimension by blurring the input image, thereby reducing perturbation artifacts along the attribution path. Dombrowski et al. smooth the network output by converting ReLUs to softplus~\cite{Dombrowski2019ExplanationsCB}.

While the above methods address noise in the attributions by manipulating the input (or the baseline), ours examines the entire path of integration. As mentioned in \cite{Sundararajan2017AxiomaticAF}, each path from the baseline to the input constitutes a different attribution method; the methods discussed above~\cite{Sundararajan2017AxiomaticAF,Smilkov2017SmoothGradRN,kapishnikov2019xrai,sturmfels2020visualizing} choose the straight line path, while ~\cite{Xu_2020_CVPR} choose the `'blur'' path when integrating gradients. 
In this work, instead of determining the path based on the input and baseline alone, we propose \emph{adaptive path methods} (APMs) that adapt the path based on the input, baseline, and the model being explained. Our intuition is that model-agnostic paths, such as the straight line, are susceptible to travel through regions that have irregular gradients, resulting in noisy attributions. We posit that adapting the integration path based on the model can avoid selecting samples from anomalous regions when determining attributions. 

We propose Guided Integrated Gradients (Guided IG), an attribution method that integrates gradients along an adaptive path determined by the input, baseline, and the model. Guided IG defines a path from the baseline towards the input, moving in the direction of features that have the lowest absolute value of partial derivatives. At each step, Guided IG selects the features (pixel intensities) with the lowest absolute value of partial derivatives (e.g., bottom 10\%), and moves only that subset closer to the intensity in the input image, leaving all others unchanged. As the intensity of specific pixels (features) becomes equal to that in the input image being explained, they are no longer candidates for selection. %
The attributions resulting from this approach are considerably less noisy. %
Experiments highlight that Guided IG outperforms %
other, related methods %
in nearly every experiment. Our main contributions are as follows:
\vspace{-4pt}
\begin{itemize}
\itemsep0em
    \item We propose Adaptive Path Methods (APMs) a generalization of path methods~\cite{Sundararajan2017AxiomaticAF} that consider the model and input when determining the attribution path.%
    \item We introduce Guided IG, an attribution technique that is an instance of an adaptive path method, and describe its theoretical properties. %
    \item We present experimental results that show Guided IG outperforms other attribution methods quantitatively and reduces noise in the final explanations.%
\end{itemize}

%% file: tex/related_works.tex
Literature on explanation and attribution methods has grown in the last few years, with a few broad categories of approaches: Black-box methods and methods that perturb the input ~\cite{ribeiro2016should,fong2017interpretable,fong2019understanding,petsiuk2018rise}; methods utilizing back-propagation~\cite{Sundararajan2017AxiomaticAF,selvaraju2017grad,BMBMS16,Bach2015OnPE}; methods that visualize intermediate layers~\cite{springenberg2014striving,Zeiler_2014,SVZ13,mahendran2015understanding,Mordvintsev2015InceptionismGD}; and techniques that combine these different approaches~\cite{kapishnikov2019xrai,Smilkov2017SmoothGradRN,bau2017network}. Our work extends and improves upon Integrated Gradients~\cite{Sundararajan2017AxiomaticAF}, a popular technique applicable to many different types of models. Accordingly, we focus on perturbation and back-propagation-based methods. %

Black-box methods such as \cite{ribeiro2016should,fong2017interpretable,fong2019understanding,petsiuk2018rise} are model agnostic, and rely on perturbing or modifying the input instance and observing the resulting changes on the model's output. While \cite{ribeiro2016should} uses segmentation-based masking of input, \cite{petsiuk2018rise} generates random smooth masks to determine salient input segments or regions. \cite{fong2017interpretable,fong2019understanding} apply multiple perturbations such as adding noise, or blurring, and optimize over the model output to learn the mask. All of these methods typically require several iterations/evaluations of the model to identify salient regions (or pixels) for a single input. As BlurIG shows~\cite{Xu_2020_CVPR}, perturbations can introduce artifacts (i.e., information not in the original image), adversely affecting the validity of the output.

Back-propagation methods~\cite{Sundararajan2017AxiomaticAF,selvaraju2017grad,BMBMS16,Bach2015OnPE} examine the gradients of the model with respect to an input instance to determine pixel-level attribution. These methods produce a saliency map by weighing the gradient contributions from layers in the network to individual pixels, or entire regions.  Our work specifically builds on the path-based Integrated Gradients in~\cite{Sundararajan2017AxiomaticAF}. Specifically, our work addresses the issue of noise in pixel attribution in IG, which is highlighted by~\cite{Smilkov2017SmoothGradRN,kapishnikov2019xrai,sturmfels2020visualizing}. While~\cite{Smilkov2017SmoothGradRN} addresses this issue by adding noise-based perturbations to the input, and averaging attributions over the perturbed input samples, \cite{kapishnikov2019xrai} aggregates attributions of regions by considering a segmentation of the input. In contrast to these previous methods, our paper addresses this issue of noise by optimizing the path along which the gradients are aggregated.

%% file: tex/gradient_section.tex
In this section, we provide a summary of the Integrated Gradients method, then describe how highly-correlated gradients can introduce noise to IG's attributions. We also show how model-agnostic paths (e.g., a straight line path) can contribute to this problem.

\subsection{Integrated Gradients}

For visual models, given an image $x$, IG calculates attributions per pixel (feature) $i$ by integrating the gradients of the function/model ($F$) output w.r.t. pixel $i$ as in Eqn.~\ref{ig_formula}.

\begin{equation}
    IG_i(x) = \int_{\alpha=0}^{1}\frac{\partial F(\gamma(\alpha))}{\partial \gamma_i(\alpha)}\frac{\partial \gamma_i(\alpha)}{\partial \alpha}d\alpha
    \label{ig_formula}
\end{equation}
where $\frac{\partial F(x)}{\partial x_i}$ is the gradient of $F$ along the $i^{th}$ feature and $\gamma(\alpha)$ represent images along some integral path ($\alpha \in [0, 1]$). 
In \cite{Sundararajan2017AxiomaticAF,Smilkov2017SmoothGradRN,kapishnikov2019xrai,miglani2020investigating}, $\gamma(\alpha)$ is a function that modifies the intensity of pixels from a baseline image $\gamma(\alpha=0)$ (e.g., a black, white, or random noise image), to that of the input being explained $\gamma(\alpha=1) = x$.

\begin{figure}
    \centering
    \begin{subfigure}[b]{0.3\linewidth}
        \centering
        \includegraphics[width=\textwidth]{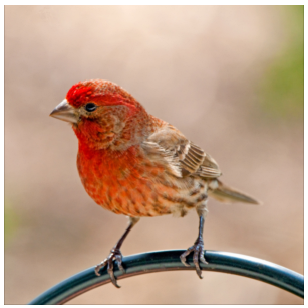}
        \caption{Input image}
        \label{fig:gradient_input}
    \end{subfigure}
    \hfill
    \begin{subfigure}[b]{0.31\linewidth}
        \centering
        \includegraphics[width=\textwidth]{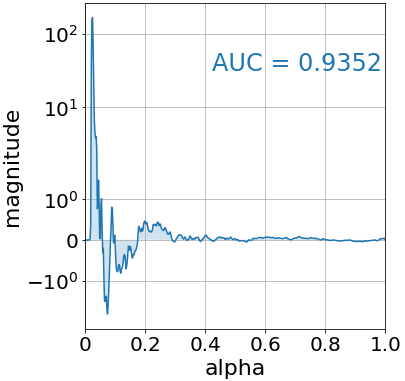}
        \caption{Directional $\Delta$}
        \label{fig:gradient_dir}
    \end{subfigure}
    \hfill
    \begin{subfigure}[b]{0.31\linewidth}
        \centering
        \includegraphics[width=\textwidth]{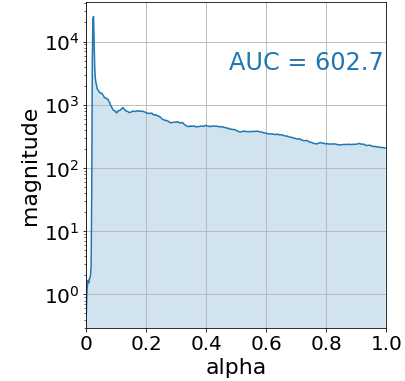}
        \caption{Gradient}
        \label{fig:gradient_magnitude}
    \end{subfigure}
      \vspace{0.2cm}
  \caption{Differences in magnitude of directional derivative and total gradients. \textbf{(a)}: The input image. \textbf{(b)}: Signed directional derivative ($\Delta$) magnitude on the straight-line path from the black baseline to the input image, using the input image from (a). The area under the curve is equal to the total attribution. \textbf{(c)}: Magnitude of gradients along the straight-line path, using the input image from (a). Even as the magnitude of directional derivatives is close to $0$ from $0.3<\alpha<1.0$, the magnitude of gradients is high, leading to possible gradient accumulation.}
  \label{fig:high-gradients}
  \vspace{-0.3cm}
\end{figure}

\subsection{Noise Originating from Model-Agnostic Paths}
\label{subsec:noise-origin}
In assessing IG's outputs, one can observe noise in the attributions (e.g., Figure \ref{fig:noise}). Recent work has investigated this noise and accredited it to the selection of baseline~\cite{sturmfels2020visualizing} or gradient accumulation in saturated regions~\cite{miglani2020investigating}.
Concurring with \cite{Dombrowski2019ExplanationsCB}, we observe that the model's surface is also an important factor.

\begin{figure}[!hbt]
  \centering
  \begin{subfigure}[b]{.85\linewidth}
  \includegraphics[width=\linewidth]{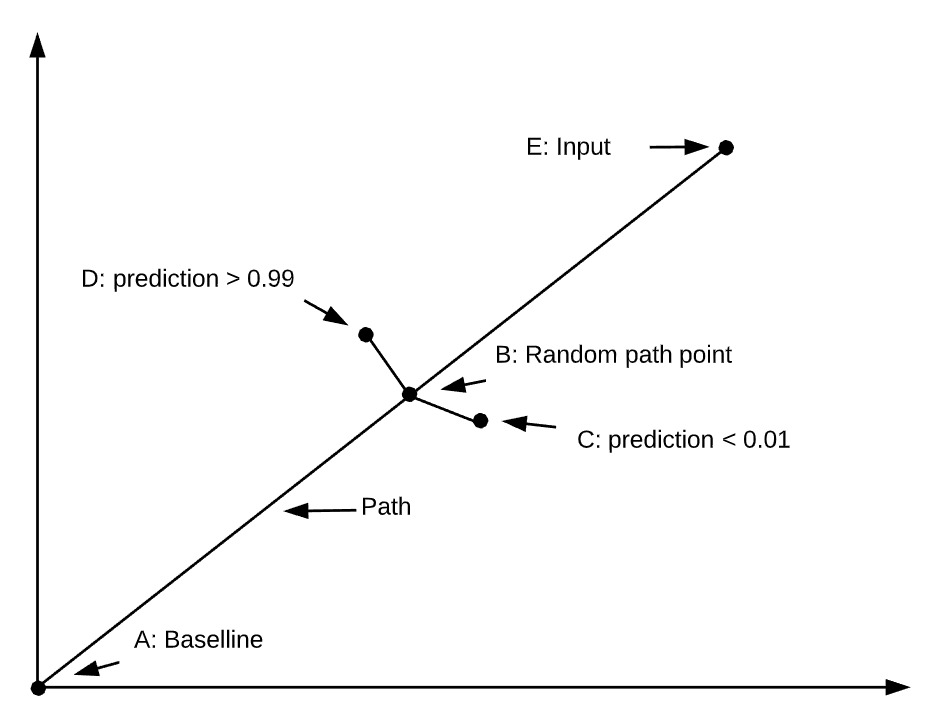}
  \label{fig:close-adversarial}
  \end{subfigure}
  \caption{High gradients along the straight-line path from baseline (A) to input (E). At any point (B) of the path, it is possible to find points (C) and (D) that are in very close proximity to (B) and have very low (C) and very high prediction scores (D) respectively. Even though these points aren’t part of the path, their close proximity to (B) implies the presence of high gradients along the straight line path.}
  \label{fig:adversarial}
\end{figure}

\begin{figure}[!hbt]
  \centering
  \includegraphics[width=1\linewidth]{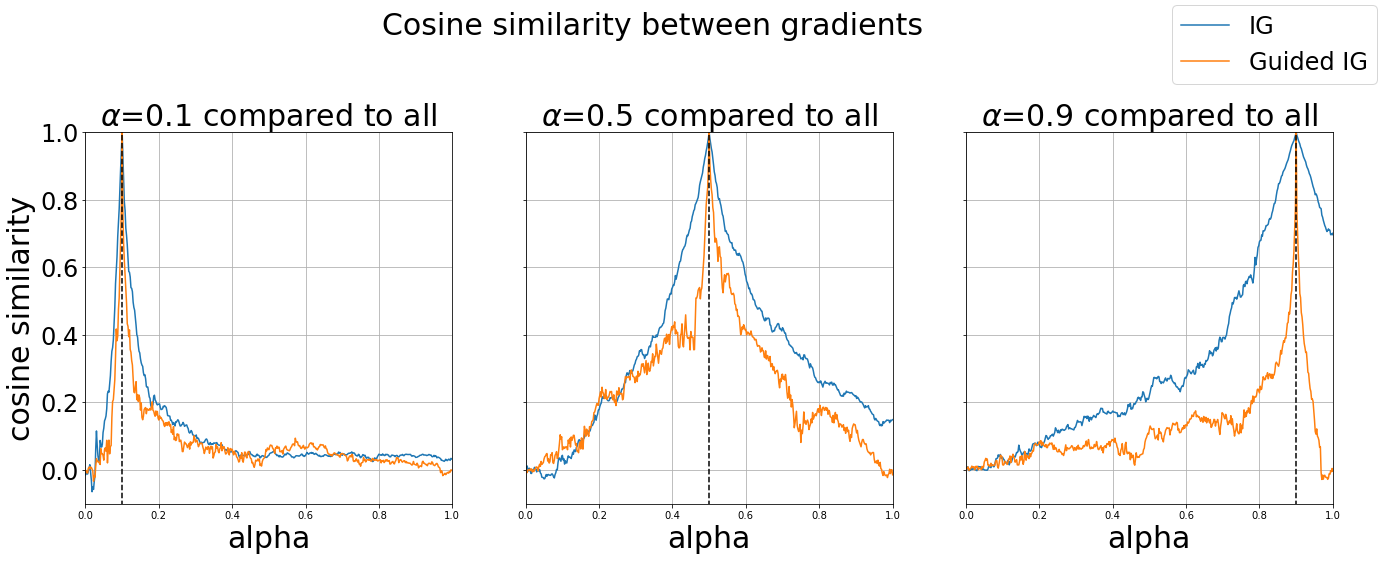}
  \caption{Correlation of gradients along the attribution path. Gradients for the Integrated Gradients (blue) and Guided IG (orange) path were calculated for the image from Figure \ref{fig:noise}. Each subplot shows cosine similarity between gradients at alpha=[0.1, 0.5, 0.9] to the gradients at all other steps of the integration path for IG and GIG. For each graph, the point chosen is indicated with a dashed vertical line.}
  \label{fig:cosine-similarity}
  \vspace{-0.3cm}
\end{figure}

Looking at the matrix of partial derivatives of the output w.r.t. the input image, we observe that the partial derivatives have a higher by-order-of-magnitude $L_2$ norm in comparison to the norm of the directional derivative (in the direction of the integration path) matrix (see Figure \ref{fig:high-gradients}). This implies that the influence of the inputs not contributing to the output may dominate the gradient map at any integral point. One would hope that gradient vectors pointing to different (incorrect) directions will cancel each other out when the whole path is integrated, but that is not always the case as gradient vectors tend to correlate (e.g., see Figure \ref{fig:cosine-similarity}). To put it plainly, spurious pixels that don't contribute to the model output end up having non-zero attributions depending on the model geometry (Figure \ref{fig:adversarial}).

A straight path, where pixel intensities are uniformly interpolated between a baseline and the input, is susceptible to travel through areas where the gradient norm is high and not pointing towards the integration path (indicated by a low cosine similarity between $\vec{\triangledown F(x)}$ and $\vec{dx}$). This issue can be minimized by averaging over multiple straight paths~\cite{Smilkov2017SmoothGradRN,kapishnikov2019xrai} or limiting the impact of noise by splitting the path into multiple segments~\cite{miglani2020investigating}. However, these approaches sidestep an important issue: attribution methods based on model-agnostic paths will highly depend on surface geometry.

%% file: tex/adaptive.tex
We introduce \textit{adaptive path methods} (APMs), a generalization of path methods (PMs), to address the limitations of model-agnostic paths.  An APM is similar to the definition of path methods (as in \cite{Sundararajan2017AxiomaticAF}), with the additional property that the path can depend on the model function. Adaptive path methods are a superset of path methods, and are defined as follows.

{\bf Definition:} Let $F:\mathbb{R}^N\rightarrow \mathbb{R}$ be a function of $X=\{x_1, ..., x_N\}$. Let $X^B=\{x_1^B, ..., x_N^B\}$ be the baseline input. Let $X^I=\{x_1^I, ..., x_N^I\}$ be the input that requires explanation. Let path $c$ be parameterized by a path function $\gamma^F=(\gamma_1^F, ..., \gamma_N^F)$ such that $x_i=\gamma_i^{F}(\alpha)$, where $\alpha\in [0, 1]$ and $\gamma_i(0)=x_i^B$ and $\gamma_i(1)=x_i^I$. The adaptive path method attribution of feature $x_i$ over curve $c$ for any input $x^I$, is defined as
\begin{equation}
a_{i}^{\gamma^F}(X^I)=\int_{\alpha=0}^1\frac{\partial F(\gamma^F(\alpha))}{\partial \gamma_i^F(\alpha)}\frac{\partial \gamma_i^F(\alpha)}{\partial \alpha}d\alpha.
\label{eq:apm_attribution}
\end{equation}

As with any path method, an APM also satisfies Implementation Invariance, Sensitivity, %
and Completeness axioms defined in Sundararajan et al.~\cite{Sundararajan2017AxiomaticAF}. Below, we expand on these properties for a specific instance of an adaptive path method, Guided IG.%

%% file: tex/guided.tex
To alleviate the effect of accumulation of attribution in directions of high gradients unrelated to the input (sec.~\ref{subsec:noise-origin}), 
we wish to define a path that avoids those (input) regions causing anomalies in the model behavior. %
We can call this ($ \ell_{noise}$), and one way to minimize over this is as follows
\begin{align}
    \gamma^{F*} &= \arg \min_{\gamma^F \in \Gamma} \ell_{noise}  \label{eq:gig_noise_loss} \\
    \ell_{noise} &= \sum_{i=1}^{N} \int_{\alpha=0}^1\left|\frac{\partial F(\gamma^F(\alpha))}{\partial \gamma_i^F(\alpha)}\frac{\partial \gamma_i^F(\alpha)}{\partial \alpha}\right|
    d\alpha \nonumber
\end{align}
By minimizing $\ell_{noise}$ at every feature (pixels $i\cdots N$)  we can hopefully avoid high gradient directions. However, before we can define $\gamma^F(\alpha)$ precisely, optimizing the above objective requires knowing the prediction surface of the neural network $F$ at every point in the input space, which is infeasible.  So, we propose a greedy approximation method called Guided Integrated Gradients.

\subsection{Guided IG}
Guided IG is an instance of an adaptive path method. As with IG, the path ($c$) starts at the baseline ($X^B$) and ends at the input being explained ($X^I$). 
However, instead of moving features (pixel intensities) in a fixed direction (all pixels incremented identically) towards the input, we make a choice at every step. At each step, we find a subset $\mathbb{S}$ of features (pixels) that have the lowest absolute value of the partial derivatives (e.g., the smallest 10\%) among those features (pixels) that are not yet equal to the input (image pixel intensity). The next step in the path is determined by moving only those pixels in $\mathbb{S}$ closer to their corresponding intensities in the input image. The path ends when all feature values (intensities of all pixels) match the input. Formally,
\begin{itemize}
  \item Let $F:R^N\rightarrow R$ be a function of $X=\{x_1, ..., x_N\}$.
\end{itemize}
The Guided IG integration path, $GIG_{(X^S, X^E, F)}$, is defined based on the starting point ($X^S$), the ending point ($X^E$), and the direction vector at every point of the curve:
\begin{equation}{\label{eq_gig_definition}}
\left[\begin{aligned}
&X^S = X^B \\
&X^E = X^I \\
&\frac{\partial \gamma_i^F(\alpha)}{\partial \alpha} = \left\{\begin{aligned}
&x_i^I - x_i^B && \text{, if $i \in \mathbb{S}$,} \\
&0 && \text{, otherwise.}
\end{aligned}
\right.
\end{aligned}
\right.
\end{equation}
$\mathbb{S}$ is calculated for every point of curve $c$, and contains features that have the lowest absolute value of the corresponding partial derivative among the features that have not yet reached the input values. More formally,
\begin{equation}\label{eq_set_s}
\mathbb{S} = \{i | \forall j : y_i \leq y_j\} \equiv \arg\min_{i}(Y)
\end{equation}
\begin{equation}\label{eq_y}
y_i =
\left\{ 
\begin{aligned}
&\left|\frac{\partial F(X)}{\partial x_i}\right| && \text{, if $i\in \{j | x_j \neq x_j^I\}$} \\
&\infty && \text{, otherwise.}
\end{aligned}
\right.
\end{equation}

\textbf{Guided IG path length.} Only features in subset $\mathbb{S}$ are changed at every point of the Guided IG path. Hence in the general case, the $L_2$ norm of the Guided IG path is greater than the norm of the straight-line path. The Cauchy--Schwarz inequality provides the upper boundary of the path length: $||GIG||_2 \leq \sqrt{N} \cdot ||IG||_2$, where $GIG$ is the Guided IG path and $IG$ is the straight line path. In terms of the $L_1$ norm, the lengths of the paths are always equal, i.e. $||GIG||_1 = ||IG||_1$. The equality is true because at every point of the Guided IG path individual features of the input are either not changed or changed in the direction of the input.

\textbf{Efficient approximation.} Guided IG can be efficiently approximated using a Riemann sum with the same asymptotic time complexity as the computation of Integrated Gradients. In domains like images, the number of features is high; therefore, it is not practical to select only one feature at every step. Hence, at every step, the approximation algorithm selects a fraction of features (we use $10\%$) with the lowest absolute gradient values and moves the selected features toward the input. %
At every step, the algorithm reduces the $L_1$ distance to the input by the value that is inversely proportional to the number of steps. The higher the number of the steps is, and the lower the fraction is, the closer the approximation is to the true value of Guided IG attribution. We provide our implementation in the Supplement. %

Since Guided IG path is computed dynamically, it is not possible to parallelize the computation for a single input. %
Parallelization is still possible when calculating attribution for multiple independent inputs (batches).

\subsection{Bounded Guided IG}

The optimal solution path to Equation \eqref{eq:gig_noise_loss} is unbounded and can deviate infinitely off the baseline-input region. However, it can be advantageous to stay close to the shortest path. First, staying close to the baseline-input region decreases the likelihood of crossing areas that are too out-of-distribution. Second, it can be computationally cheaper to numerically approximate the integral of a shorter path.

To this end, we modify the objective in Eqn.~\ref{eq:gig_noise_loss} to additionally minimize the accumulated distance to the straight-line path ($\ell_{distance}$). The new objective can then be defined as:
\begin{align}
    \gamma^{F*} &= \arg \min_{\gamma^F \in \Gamma} \ell_{noise} + \lambda \ell_{distance} \label{eq:gig_optimal_loss} \\
    \ell_{distance} &= \int_{\alpha=0}^1\left|\left|\gamma^F(\alpha)-\gamma^{IG}(\alpha)\right|\right|d\alpha \nonumber,
\end{align}
where $\gamma^{IG}(\alpha)$ is the parameterized straight-line path and $\lambda$ is the coefficient that balances the two components. For very large values of $\lambda$ (e.g. $\lambda=\infty$), the solution of this objective reduces to the shortest path (same as the IG path). Setting  $\lambda = 0$, would give us Eqn.~\ref{eq:gig_noise_loss}, which can be thought of as an \textit{unbounded} version.

One approximation of this objective is limiting the maximum distance that the Guided IG path can deviate from the straight line path at any point\footnote {There are multiple ways of limiting the distance. See \url{https://github.com/pair-code/saliency} for the latest implementation.}.
We introduce the concept of \textit{anchors} as a simple way of achieving this. We divide the straight-line path between the baseline $X^B$ and the input $X^I$ into $K+1$ segments and compute Guided IG for each segment separately, effectively forcing the Guided IG path to intersect with the shortest path at $K$ anchor locations. We call this the anchored Guided IG.
Accordingly, selecting a higher number of anchors corresponds to optimizing for a higher value of $\lambda$ in Equation \ref{eq:gig_optimal_loss}, making the results of Guided IG closer to IG.
On the other hand, when the number of anchors is zero we have the unbounded algorithm described previously. To put it concretely,
for the $k^{th}$ segment, its starting ($X^S_k$) and ending ($X^E_k$) point can be set as
\begin{itemize}
\item $X^S_k = (X^I - X^B) (k-1)/(K+1) + X^B$, and
\item $X^E_k = (X^I - X^B) (k)/(K+1) + X^B$
\end{itemize}
We then define Guided IG with $K$ anchors as: 
\begin{equation*}
GIG_{(X^S, X^E, F)}(K) = \sum_{k=1}^{K+1} GIG_{(X^S_k, X^E_k, F)}.
\end{equation*}

Since the integral is a linear operator, summing the integrals of each individual segment is the same as taking the integral of the whole path. For simplicity, we will ignore the other terms and refer to Guided IG with $K$ anchor points as $GIG(K)$ in the rest of the paper.

Sometimes, having zero anchors is more favorable as there can be cases where anchors overlap with the high gradient regions on the shortest path. The more anchors there are, the more likely it is to hit a high gradient region and accumulate noise, while staying closer to the shortest path. We will show the effect of anchors in the Results section in more detail.

\subsection{Axiomatic Properties of Guided IG}
Guided IG satisfies a subset of desired axioms as IG\cite{Sundararajan2017AxiomaticAF}. We note the ones we satisfy below. As with any path integral in a conservative vector field, Guided IG satisfies the completeness axiom that can be summarized by Eq. \eqref{eq:completeness} (where $a_i$ denotes the attribution per pixel/feature)
\begin{equation}\label{eq:completeness}
\sum_{i=1}^{N}a_i=F(X^I)-F(X^B)
\end{equation}
Since Guided IG satisfies completeness, it also satisfies Sensitivity(a) (see \cite{Sundararajan2017AxiomaticAF} for the proof). %
Sensitivity(b)\cite{Friedman2004PathsAC} is also satisfied because the partial derivative of a function with respect to a dummy variable is always zero at any point of the path. As a result, the value of the integral in Eq. \eqref{eq:apm_attribution} is always zero for any such variable.

In the appendix, we provide a proof that Guided IG is Symmetry-Preserving. Symmetry guarantees that if $F(x, y) = F(y, x)$ for all $x$ and $y$ then both $x$ and $y$ should always be assigned equal attribution. The important remark here is that this does not contradict the uniqueness of the IG method. IG satisfies Symmetry for any function; however, given a function, it may be possible to find other paths that satisfy Symmetry. In practice, we always calculate attributions for a given function, e.g., a neural network model.

Guided IG satisfies the Implementation Invariance axiom; thus, it always produces identical attributions for two functionally equivalent networks. Guided IG preserves the invariance since it only relies on the function gradients and does not depend on the internal structure of the network.  There may be other properties that are worth further study such as additivity and uniqueness, please refer to ~\cite{sundararajan2020many}.

%% file: tex/results.tex
We evaluate Guided IG by first observing its behavior in attributing a closed path, then examine its performance using common benchmarks, datasets, and models.

\subsection{Attribution of a Closed Path}
One challenge in evaluating attribution methods is the lack of ground truth for the attributions themselves~\cite{miglani2020investigating}.
The Completeness axiom guarantees that the sum of all feature attributions for any path is equal to the difference between the function value at the input and the function value at the baseline. Therefore, the sum of all attributions for a closed path is equal to zero. However, Completeness does not define the values of individual feature attributions, only the sum.
We can, however, axiomatically define the ground truth attribution for all features as the average attribution values of all possible paths from the baseline to the input. This definition is similar to Shapley values \cite{shapley1953value} that are also defined in terms of a sum of all possible paths.
Using the axiomatic definition of the ground truth attribution, we can now prove that the ground truth attribution for a closed path $A\rightarrow A$ is zero for all features.

\textbf{Proof}. Let $P$ be a set of all possible paths from point $A$ to point $A$. Let $a$ denote attribution. For any path $p\in P$, there exists a counterpart reverse path $p'$ such that $a_i(p)+a_i(p')=0$ for all features $i$. Since every path has a counterpart reverse path cancelling its attributions, the average attribution values of individual features are $0$. Q.E.D.

\begin{figure}
  \centering
  \includegraphics[width=0.9\linewidth]{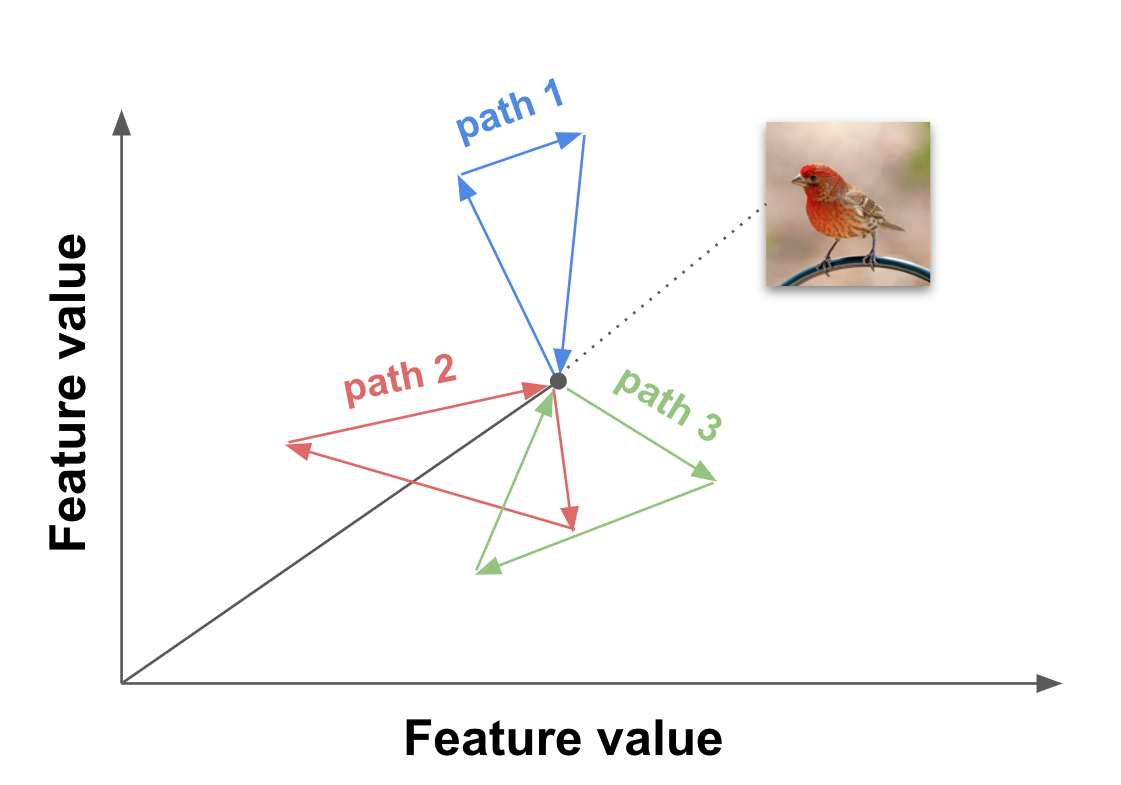}
  \caption{Attribution of closed paths. By calculating the attribution path of the input image with itself via random points, we create an attribution path that in theory should be zero.}
  \label{fig:triangle-path}
\end{figure}

We can build a random path $A\rightarrow B \rightarrow C \rightarrow A$ and consider it as an estimator of ground truth attribution of path $A\rightarrow A$. Using the path integral additive property, we can break the path into sub-segments, i.e., $a(A\rightarrow B \rightarrow C \rightarrow A) = a(A\rightarrow B) + a(B\rightarrow C) + a(C \rightarrow A)$. Now, we can apply a path method on every segment and treat the sum as an estimation of the ground truth. Figure \ref{fig:triangle-path} gives an illustration of the idea.

We apply this technique to calculate the attribution of each segment using both IG and Guided IG. We sample 50 random paths on 200 random images from the ImageNet validation dataset, for the total of 10,000 path samples. We calculate the mean of the squared error for individual features and average the error across all images, pixels, and channels. The results in Table~\ref{table:closed} show that applying Guided IG results in lower error compared to IG.
\vspace{-0.2cm}
\begin{table}[!ht]
\centering
\resizebox{0.45\textwidth}{!}{
\begin{tabular}{ c|c|c|c } 
 & MobileNet & Inception & ResNet \\ \hline
\makecell  {IG} & 3.817e-07  & 5.938e-07 & 6.707e-07 \\ \hline 
\makecell  Guided IG & \textbf{7.320e-08}  & \textbf{1.442e-07} &  \textbf{1.857e-07} \\ \hline 
\end{tabular}}
\caption{Closed path mean squared error for IG and Guided IG.}
\label{table:closed}
\end{table}

\begin{table}[!bht]
\footnotesize
\setlength{\tabcolsep}{2pt}
\begin{center}
\resizebox{0.7\columnwidth}{!}{\begin{tabular}{l|ccc|c|c}
\toprule
\multicolumn{1}{c|}{{(AUC)}} & \multicolumn{3}{c}{{ImageNet}} & \multicolumn{1}{|c}{{Open Images}} & \multicolumn{1}{|c}{{DR}} \\
\multicolumn{1}{c|}{{Method}} & MobileNet & Inception & ResNet & ResNet & ~\cite{krause_2018} \\
\midrule
Edge &   0.611 &   0.610 &  0.611 & 0.606 & 0.643 \\
Gradients & 0.614 &  0.634 &  0.650 & 0.505 &  0.801 \\
IG & 0.629 &  0.655  &  0.669 & 0.557 & 0.833 \\
Blur IG &  0.652 &  0.662 & 0.663 & 0.619 & 0.830 \\
GIG (0) & \textbf{0.705} &  \textbf{0.712} & \textbf{0.711} & \textbf{0.630} & 0.619 \\
GIG (20) &  0.691 & 0.696 & 0.706 &  0.624 &  \textbf{0.863}$^*$ \\
\midrule
GradCAM &   0.776 &   0.761 &  0.755 & 0.474 & 0.837 \\
\midrule
Smoothgrad \\
+IG &  0.742 & 0.773   & 0.781	  & 0.662   & 0.637  \\
+GIG(0) &  0.745 & 	0.776	 & 0.776 & 0.649 & 0.632 \\
+GIG(20) &  \textbf{0.767} & \textbf{0.795} & \textbf{0.799}	 & \textbf{0.685} & \textbf{0.645} \\
\midrule
XRAI \\
 +IG &  0.731	 & 0.765   & 0.762  & 0.631 & 0.793 \\
 +GIG(0) & \textbf{0.838}$^*$	 & \textbf{0.829}$^*$ & \textbf{0.821}$^*$ & 0.718 & 0.630  \\
 +GIG(20) &  0.808 & 0.819 & 0.809 & \textbf{0.719}$^*$ & \textbf{0.831} \\
\bottomrule
\end{tabular}}
\caption[qnt-eval]{
We report the mean AUC values for different methods using the black baseline for the methods compared. Higher is better. \textbf{bold} indicates highest in each group, * indicates highest overall.
}\label{tab:qnt-eval}
\end{center}
\vspace{-0.3cm}
\end{table}
\vspace{-0.3cm}
\subsection{Quantitative evaluation}
\paragraph{Metrics}
We compare Guided IG attributions with other attribution methods by employing the \textbf{AUC-ROC} metric as described in \cite{BylinskiiJOTD16}. The metric treats the attributions as classifier prediction scores. Ground truth is provided by human annotators. The sliding threshold determines the proportion of features that are assigned to the ``true'' class. By changing the threshold, the ROC curve is drawn and the AUC of that curve is calculated.

Additionally, we also use the \textbf{Softmax Information Curve (SIC AUC)} metric from \cite{kapishnikov2019xrai}. This method directly measures how well the model performs without using human evaluation. It does so by revealing only the regions that are highlighted by the attribution method and measuring the model's softmax score. The key idea is that the attribution method that has better focus on where the model is truly looking should reach the softmax value faster than another one that is less focused on the correct region. Therefore, this metric evaluates the attribution method from the model's perspective without any human involvement.

\begin{table}[!tb]
\setlength{\tabcolsep}{2pt}
\centering
\small
\resizebox{0.9\columnwidth}{!}{\begin{tabular}{l|ccc|ccc|ccc}
\toprule
\multicolumn{1}{c|}{{(AUC)}} & \multicolumn{3}{c|}{{ImageNet-Inception}} & \multicolumn{3}{c|}{{OpenImages}} & \multicolumn{3}{c}{{DR}} \\
\multicolumn{1}{c|}{{Method}} & black & b+w & 2-rnd & black & b+w & 2-rnd &  black & b+w & 2-rnd \\
\midrule
IG        & 0.655 & 0.667 & 0.689 & 0.557 & 0.572 & 0.600 & 0.833 & 0.824  & 0.791 \\
Blur IG   & 0.662 & 0.662 & 0.662 & 0.619 & 0.619 & \textbf{0.619} & 0.830 & 0.830 & \textbf{0.830} \\
GIG(0)    & 0.712 & 0.738 & 0.722 & 0.630 & 0.625 & 0.607 & 0.619 & 0.544 &  0.510\\
\midrule
GIG(10) & 0.702 & 0.722 & 0.709 & 0.626 & 0.636 & 0.617 & 0.850 & 0.837 & 0.751 \\
GIG(20)   & 0.696 & 0.714 & 0.704 & 0.624 & 0.639 & 0.617 & 0.863 & 0.850  & 0.792 \\
GIG(40) & 0.690 & 0.706 & 0.698 & 0.615 & 0.634 & 0.613 & \textbf{0.865}$^*$ & \textbf{0.851} & 0.794 \\
\midrule
XRAI \\
+IG & 0.765   & 0.820 & 0.843 & 0.631 & 0.697	 & 0.747 & 0.793 & 0.810 & 0.793                          \\
+GIG(0) & \textbf{0.829} & \textbf{0.854}$^*$ & 0.843 & \textbf{0.718} & 0.714 & 0.674	 & 0.630 & 0.629 & 0.573 \\
+GIG(20) & 0.819   & 0.852& \textbf{0.851}& 0.719 & 0.764 & 	0.744 & 0.831 & 0.831 & 0.788                          \\
+GIG(40) & 0.815	 & 0.852 & 0.849 & 0.709 & \textbf{0.768}$^*$ & \textbf{0.749} & 0.829	 & 0.831 & 0.795 \\
\bottomrule
\end{tabular}}
\caption{AUC-ROC values for ImageNet Inception model, the Open Images ResNet model, and the DR model, using 3 different choices of baseline - black, black and white, and 2 random (note that BlurIG does not need a baseline); and 4 different anchored versions of GIG (K=\{0, 10, 20, 40\}). Higher is better, values in \textbf{bold} are highest in each column, $^*$ is highest on dataset. \label{tab:baseline_anchors_choice}}
\vspace{-0.3cm}
\end{table}

\begin{table}[!htb]
\setlength{\tabcolsep}{2pt}
\centering
\small
\resizebox{0.7\columnwidth}{!}{\begin{tabular}{@{}l|ccc|c@{}}
\toprule
    (SIC AUC)      & \multicolumn{3}{c|}{ImageNet}                    & \multicolumn{1}{l}{Open Images} \\
\multicolumn{1}{c|}{\begin{tabular}[c]{@{}c@{}} Method\end{tabular}} & MobileNet & Inception & ResNet & ResNet \\ \midrule
Edge      & 0.300          & 0.371          & 0.405          & 0.537                           \\
Gradients & 0.368          & 0.431          & 0.510          & 0.595                           \\
IG        & 0.402          & 0.499          & 0.544          & 0.694                          \\
Blur IG   & 0.411          & 0.501          & .560          & 0.659                           \\
GIG(0)    & 0.423          & 0.516         & 0.550          & 0.634                           \\
GIG(20)   &      0.453          & 0.546          & 0.584          & 0.701                           \\
GIG(40)   & 0.453 & 0.551 & 0.592 & 0.734 \\
\midrule
GradCAM & 0.691 & 0.739 & 0.763  & 0.662 \\
\midrule
XRAI \\
 +IG & 0.671 & 0.736 & 0.755  &  0.843 \\
 +GIG(40) & \textbf{0.692} &\textbf{ 0.752} & \textbf{0.771} & \textbf{0.866}
\\ \bottomrule
\end{tabular}}
\caption{SIC AUC results \cite{kapishnikov2019xrai} for Guided IG and other methods. All methods use black and white baseline if they require it. Values in \textbf{bold} are the highest. 
\label{tab:sic_results}}
\vspace{-0.3cm}
\end{table}

\textbf{Methods} 
We compare our method against four baselines: edge detector, vanilla Gradients \cite{SVZ13}, IG, and Blur IG \cite{Xu_2020_CVPR} (with max $\sigma=35$). We compare both the unbounded version of Guided IG (listed as GIG(0), where the number of anchors is in parentheses), and variants with anchors (e.g., GIG(20)). The edge detector saliency for an individual pixel is calculated as the average absolute difference between the intensity of the pixel and the intensity of its nearest eight adjacent neighbours. We used 200 steps and the black baseline for all methods that required these parameters. 
\vspace{-0.35cm}
\subsubsection{Datasets}
\vspace{-0.15cm}
We evaluated our approach on two datasets of natural images, and one dataset of medical images (described below), and report results in Table~\ref{tab:qnt-eval}.

\textbf{ImageNet}~\cite{russakovsky_imagenet_2014} 
We used images from the standard validation set. We only included images that had ground truth annotation and were predicted as one of the top 5 classes by the corresponding model. We calculated the AUC-ROC metric for the ImageNet dataset using three pre-trained models: Mobilenet\_v2 (n=4016), Inception\_v2 (n=3965) and Resnet\_v2 (n=3838).\footnote{All models were downloaded from TensorFlow Hub.}

\textbf{Open Images}~\cite{Kuznetsova_2020} 
We evaluated on 5000 random images from the validation set of the Open Images dataset. As with ImageNet, we only included images that had ground truth annotation and were predicted as one of the top 5 classes by the corresponding model.

\textbf{Medical Images}~\cite{eyepacs_2009}
We also compare our method on a model trained to predict Diabetic Retinopathy (DR). Specifically, we use the Inception-v4 DR classification model from \cite{krause_2018}. We examine the results on a sample of 165 images from the validation set~\cite{eyepacs_2009}. %

\begin{figure}
    \centering
    \begin{subfigure}[b]{0.45\linewidth}
        \centering
        \includegraphics[width=\textwidth]{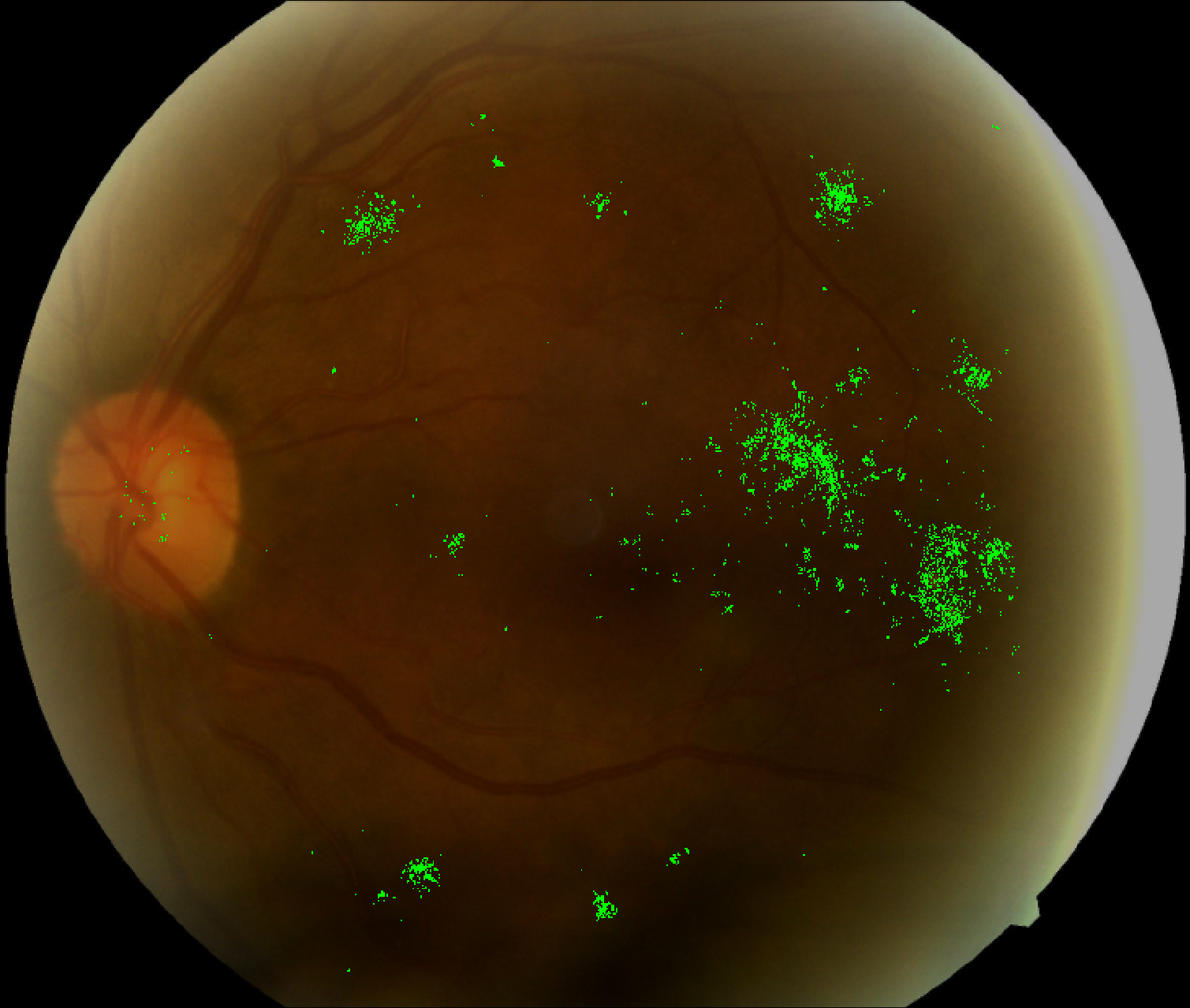}
        \caption{IG attribution}
        \label{fig:dr_ig}
    \end{subfigure}
    \hfill
    \begin{subfigure}[b]{0.45\linewidth}
        \centering
        \includegraphics[width=\textwidth]{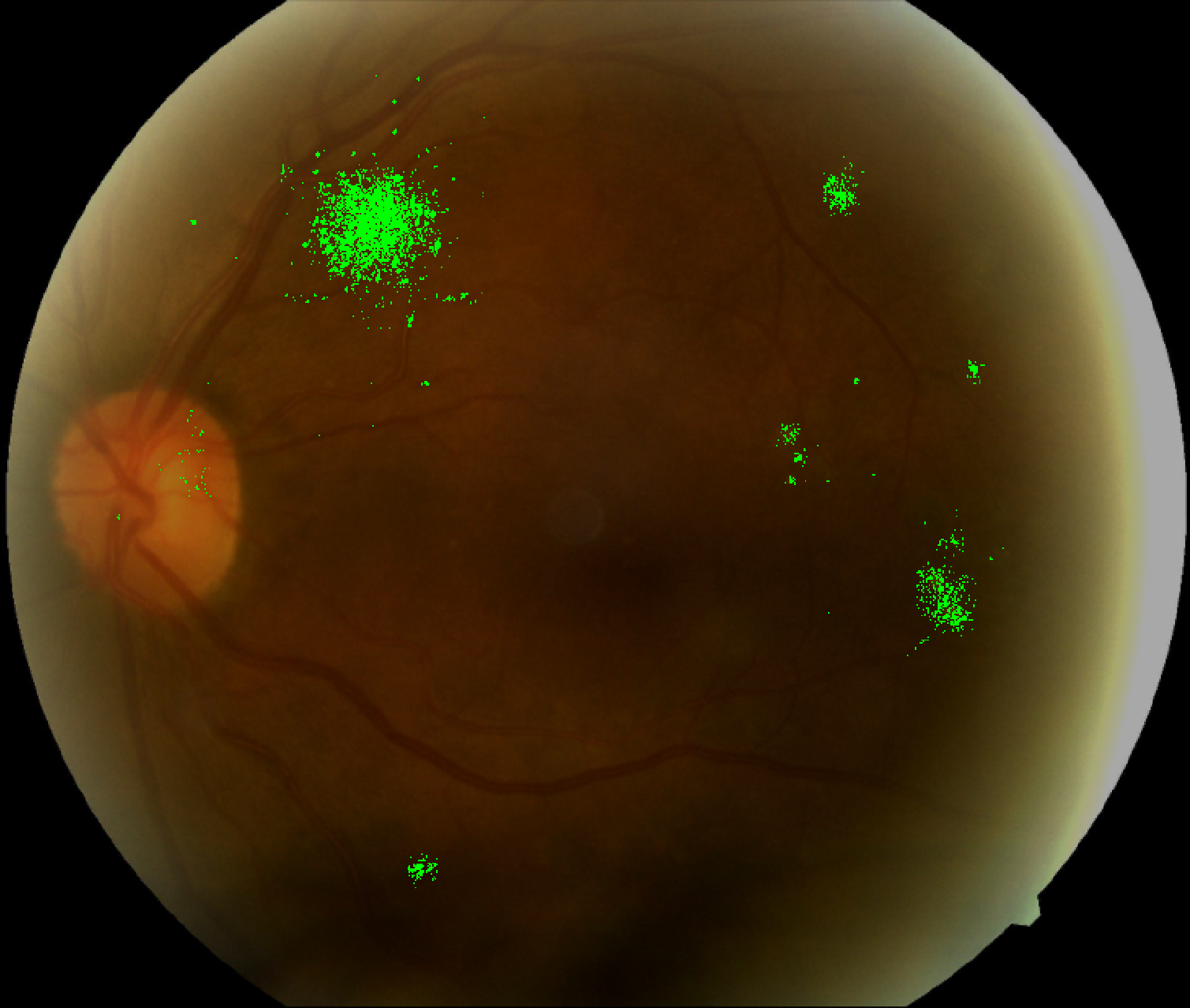}
        \caption{Guided IG attribution}
        \label{fig:dr_gig}
    \end{subfigure}
  \caption{Comparison of IG and Guided IG on a retina image~\cite{eyepacs_2009} used in diagnosing diabetic retinopathy.
}
\vspace{-0.3cm}
\end{figure}

From Tables~\ref{tab:qnt-eval} and ~\ref{tab:sic_results}, we can see Guided IG outperforms other methods. Also, Smoothgrad~\cite{Smilkov2017SmoothGradRN} and Guided IG are likely reducing the same sources of noise, so we only see a marginal improvement when combining the two methods. However, adding anchors (e.g., GIG(20)), shows a substantial improvement in performance. We also note that smoothing does not seem to be a good strategy on the DR dataset where sparser attributions may be preferred; this can also be observed with XRAI, but to a lesser extent. The XRAI method aggregates attributions to image segments, and hence when combined with Guided IG, it shows the best performance on most models. 
\vspace{-0.41cm}
\subsubsection{Effect of baseline choice}
\vspace{-0.28cm}
For path methods, there are different options one can choose for the baseline. Table~\ref{tab:baseline_anchors_choice} examines the effect of choosing a black baseline, (average over a) black and a white baseline, and (the average over) 2 random baselines. While a black+white baseline is generally a better choice, with GIG(0), we can see that much of the improvement (over IG) is observed on a single black baseline itself. 
\vspace{-0.35cm}
\subsubsection{Effect of number of anchors}
\vspace{-0.28cm}
We also report the performance of anchored versions of Guided IG. Table~\ref{tab:baseline_anchors_choice} examines all the models on 4 different choice of anchors $K=\{0, 10, 20, 40\}$ (where $0$ is simply the unbounded version of Guided IG.) For natural images, it appears that Guided IG without any anchors is a better choice. On the DR dataset, Guided IG seems reliant on anchor points along the straight line path.
Overall, Guided IG with a black or black+white baseline and 20 anchor points leads to consistently good performance on the evaluated metrics across all models and datasets.

\vspace{-3pt}
\subsection{Qualitative Results}
\vspace{-3pt}
Figure \ref{fig:qualitative_in_paper} shows a sampling of results for IG and Guided IG (more in Appendix)
with Inception v2 as the model. As can be seen in these figures, Guided IG generally clusters its attributions around the predicted class object with comparatively less noise in other areas of the image. 
We provide additional qualitative results including failure examples, proof of symmetry, and pseudocode in the Supplement.
\begin{figure}[!thb]
    \centering
    \includegraphics[width=0.85\linewidth]{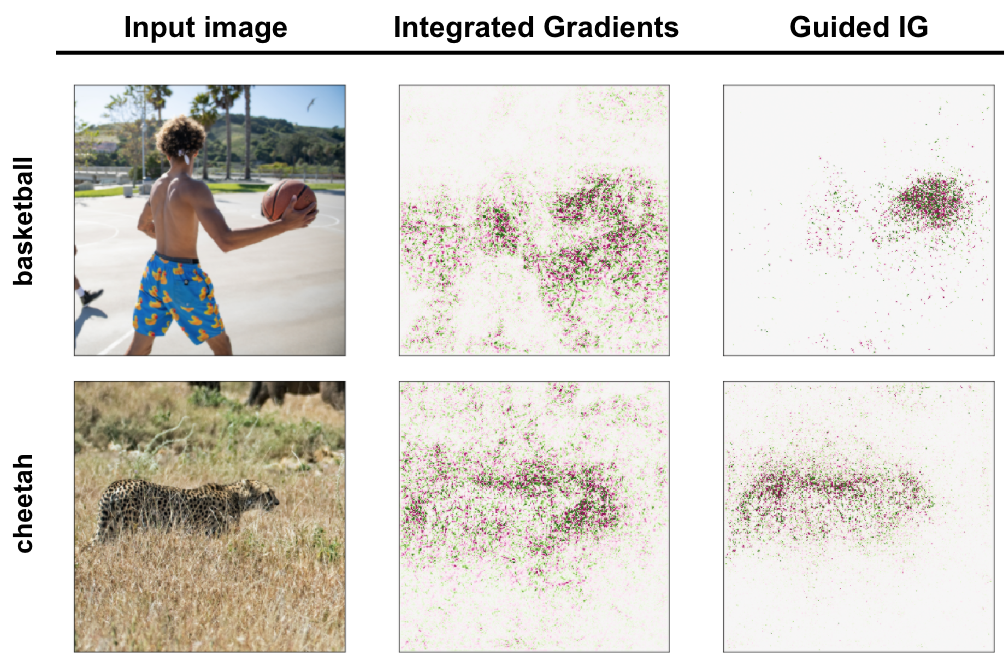}
  \caption{Visual comparison of GIG and IG. See how GIG is more focused around the basketball compared to IG.     \label{fig:qualitative_in_paper}}
  \vspace{-0.3cm}
\end{figure}

%% file: tex/discussion.tex
Experimental results from Tables~\ref{tab:qnt-eval} and~\ref{tab:baseline_anchors_choice} show that adapting the path to avoid high gradient information allows Guided IG to perform better than other tested methods. From Table~\ref{tab:baseline_anchors_choice}, we can also observe that Guided IG performs well irrespective of the choice of baseline. 

In the experimental results, we observe some variation in Guided IG's performance as a function of the number of anchor points. In many cases, these differences are relatively minor; from our experimental results, 20 anchor points may be a reasonable default value to use for this technique. Additionally, while our implementation employs anchor points to prevent paths from becoming too long, there are other ways one could achieve this goal. For example, one could ``bound'' the path to prevent it from straying too far from the straight line path of IG. Future work could explore alternative methods like this to ensure adaptive paths reduce accumulation of noise, without sacrificing either attribution quality or performance.

Guided IG demonstrates only one instance of an APM; there may exist other APM instances that are better suited for particular tasks, domains, models. Moreover, while we evaluated Guided IG on visual models and datasets, the same or different variants may be better suited for other modalities such as text or graph models.

%% file: tex/conclusion.tex
This paper introduces the concept of Adaptive Path Methods (APMs) as an alternative to straight line paths in Integrated Gradients. We motivate APMs by observing how attribution noise can accumulate along a straight line path. APMs form the basis for Guided IG, a technique that builds on IG and 
adapts the path of integration to avoid introduction of attribution noise along the path, while still optionally minimizing the path length to back-off to a straight path.
We demonstrate that Guided IG achieves improved results on common attribution metrics for image models. Opportunities for future work include understanding and  investigating Guided IG on other modalities, such as text or graph models.

%% file: tex/appendix.tex
\input{tex/symmetry}
\newpage
\input{tex/pseudocode}
\input{tex/galleries}

%% file: tex/symmetry.tex
\section{Proof of Symmetry}
\subsection{Lemma 1}
Let $F(x_1, x_2, x_3, ..., x_N)$ be a function of $N$ variables: $\mathbb{R}^N\rightarrow \mathbb{R}$.
Without loss of generality, let $x_1$ and $x_2$ be symmetric variables, i.e.
\begin{equation}\label{eq_sym_var}
\begin{split}
&F(x_1, x_2, x_3, ..., x_N) = F(x_2, x_1, x_3, ..., x_N), \\
& \text{for } \forall x_1, x_2\in \mathbb{R}.
\end{split}
\end{equation}

{\bf Claim:}
\begin{equation}
\begin{split}
&\frac{\partial F}{\partial x_1}(x_1, x_2, x_3, ..., x_N)=\frac{\partial F}{\partial x_2}(x_1, x_2, x_3, ..., x_N), \\
& \text{for } \:\forall x_1, x_2\in \mathbb{R}, x_1 = x_2
\end{split}
\end{equation}

In other words, the partial derivatives of a function with respect to symmetric variables are equal if the values of the symmetric variables are equal.

{\bf Proof:}  

Let $x_1=x_2=a$. From the definition of partial derivative:
\begin{equation}\label{eq_der_proof}
\begin{aligned}
&\frac{\partial F}{\partial x_1}(a, a, x_3, ..., x_N) \\
&= \lim_{h\rightarrow 0} \frac{F(a + h, a, x_3, ..., x_N)-F(a, a, x_3, ..., x_N)}{h}  \\
&= \lim_{h\rightarrow 0} \frac{F(a, a + h, x_3, ..., x_N)-F(a, a, x_3, ..., x_N))}{h}\text{ by \eqref{eq_sym_var}} \\
&= \frac{\partial f}{\partial x_2}(a, a, x_3, ..., x_N)\text{ Q.E.D.}
\end{aligned}
\end{equation}
\subsection{Lemma 2}
{\bf Definition:} An attribution method is symmetry preserving if, for all inputs that have identical values for symmetric variables and baselines that have identical values for symmetric variables, the symmetric variables receive identical attributions.

{\bf Claim:}
\begin{equation}
a_{i}^{\gamma^F}(X^I) = a_{j}^{\gamma^F}(X^I) \text{ if } x_i=x_j \text{ for } \forall\alpha\in[0, 1].   
\end{equation}

In other words, if the values of symmetric variables are equal at every point of the integration path then their attributions are equal. Therefore, such a path is symmetry preserving. 

{\bf Proof:} The path method attribution of variable $x_i$ is defined by \eqref{eq:apm_attribution} and is repeated here:
\begin{equation}
a_{i}^{\gamma^F}(X^I)=\int_{\alpha=0}^1\frac{\partial F(\gamma^F(\alpha))}{\partial \gamma_i^F(\alpha)}\frac{\partial \gamma_i^F(\alpha)}{\partial \alpha}d\alpha.
\label{eq:apm_attribution_appendix}
\end{equation}
Since $x_i=\gamma_i^F(\alpha)$  and $X=\gamma^F(\alpha)$ by definition, let's simplify the notations in \eqref{eq:apm_attribution_appendix} as

\begin{equation}
a_{i}^{\gamma^F}(X^I)=\int_{\alpha=0}^1\frac{\partial F(X)}{\partial x_i}\frac{\partial x_i}{\partial \alpha}d\alpha.
\end{equation}
Let $x_i$ and $x_j$ be symmetric variables for which the symmetry should be proved. We need to prove that their attribution is equal, i.e.
\begin{equation}
\int_{\alpha=0}^1\frac{\partial F(X)}{\partial x_i}\frac{\partial x_i}{\partial \alpha}d\alpha=\int_{\alpha=0}^1\frac{\partial F(X)}{\partial x_j}\frac{\partial x_j}{\partial \alpha}d\alpha
\end{equation}
\begin{equation}
\int_{\alpha=0}^1\frac{\partial F(X)}{\partial x_i}\frac{\partial x_i}{\partial \alpha}d\alpha - \int_{\alpha=0}^1\frac{\partial F(X)}{\partial x_j}\frac{\partial x_j}{\partial \alpha}d\alpha = 0
\end{equation}
\begin{equation}\label{eq_integ_is_zero}
\int_{\alpha=0}^1\frac{\partial F(X)}{\partial x_i}\frac{\partial x_i}{\partial \alpha} - \frac{\partial F(X)}{\partial x_j}\frac{\partial x_j}{\partial \alpha}d\alpha = 0
\end{equation}

To prove \eqref{eq_integ_is_zero}, it is sufficient to prove that
\begin{equation}
\frac{\partial F(X)}{\partial x_i}\frac{\partial x_i}{\partial \alpha} - \frac{\partial F(X)}{\partial x_j}\frac{\partial x_j}{\partial \alpha} = 0 \text{, for } \forall\alpha\in[0, 1]
\end{equation}

According to the premise of the claim, $x_i = x_j$ for all $\alpha$. This can only be true if the rate of change of $x_i$ is equal to the rate of change of $x_j$ for all $\alpha$, i.e., $\frac{\partial x_i}{\partial \alpha}=\frac{\partial x_j}{\partial \alpha}$. By substituting $\frac{\partial x_j}{\partial \alpha}$ with $\frac{\partial x_i}{\partial \alpha}$,
\begin{equation}
\frac{\partial F(X)}{\partial x_i}\frac{\partial x_i}{\partial \alpha} - \frac{\partial F(X)}{\partial x_j}\frac{\partial x_i}{\partial \alpha} = 0
\end{equation}
According to \emph{Lemma 1}, $\frac{\partial F(X)}{\partial x_i} = \frac{\partial F(X)}{\partial x_j}$ for all $x_1 = x_2$ if the variables are symmetric. By substituting $\frac{\partial F(X)}{\partial x_j}$ with $\frac{\partial F(X)}{\partial x_i}$, we get
\begin{equation}
\frac{\partial F(X)}{\partial x_i}\frac{\partial x_i}{\partial \alpha} - \frac{\partial F(X)}{\partial x_i}\frac{\partial x_i}{\partial \alpha} = 0,\; \text{Q.E.D}.
\end{equation}

\subsection{Lemma 3}
Let $x_i$ and $x_j$ be the values of symmetric features for which symmetry preservation should be proven. Let $\mathbb{S}$ be a set as defined by \eqref{eq_set_s}.

{\bf Claim:}
\begin{equation}
\{i, j\} \subset \mathbb{S} \text{ or } \{i, j\} \cap \mathbb{S} = \emptyset, \text{ for } x_i = x_j.
\end{equation}
In other words, if at a point of the Guided IG path $x_i$ = $x_j$, then features $i$ and $j$ are either both included in $\mathbb{S}$ or both excluded from $\mathbb{S}$ at that point.

{\bf Proof:} Due to properties of the $argmin$ function \eqref{eq_set_s}, features $i$ and $j$ are either both included or both excluded if $y_i = y_j$ (see \eqref{eq_y}). Thus, it is sufficient to prove that $y_i = y_j$ if $x_i = x_j$. The symmetry preservation is only defined for features that have the same value at the input; therefore, if $x_i$ = $x_j$ then either $(x_i = x_j) = (x_i^I = x_j^I)$ or $(x_i = x_j) \neq (x_i^I = x_j^I)$. For the former case, $y_i = y_j = \infty$. For the latter case, it is necessary to prove that the partial derivative of $x_i$ and $x_j$ are equal for symmetric variables if $x_i = x_j$, which is proved by \emph{Lemma 1}. Thus, $y_i = y_j$ is always true if $x_i = x_j$, Q.E.D.

\subsection{Theorem 1}

{\bf Claim:} The guided integrated gradients method is symmetry preserving.

{\bf Proof:} 
According to \emph{Lemma 2}, a path method is symmetry preserving if the values of symmetric variables are equal at every point of the integration path. For any two variables to have same values at every point of the path, it is sufficient for them to:
\begin{enumerate}
\item have equal values at the starting point of the path, and
\item have equal rates of change if the the values of the variables are equal.
\end{enumerate}
The proof of these two sufficiency requirements follows next.

To prove the starting point equality, let $x_i$ and $x_j$ be two symmetric variables for which the sufficiency requirement has to be proven.
The symmetry preservation is defined only for symmetric variables that are equal at the input and are equal at the baseline, i.e.,
\begin{equation}\label{eq_sym_val_at_start}
x_i^I = x_j^I \text{ and } x_i^B = x_j^B.
\end{equation}
From the Guided IG definition \eqref{eq_gig_definition},
\begin{equation}\label{eq_gig_val_at_start}
x_i^S = x_i^B \text{ and } x_j^S = x_j^B.
\end{equation}
From \eqref{eq_sym_val_at_start} and \eqref{eq_gig_val_at_start},
\begin{equation}\label{eq_req_1}
x_i^S = x_i^I = x_j^I = x_j^S,\; \text{Q.E.D. for req. 1.}
\end{equation}

To prove that the rate of change is always equal, let's assume that $x_i = x_j$ and consider two possible scenarios according to \emph{Lemma 3}: $\{i, j\} \subset \mathbb{S}$ and $\{i, j\} \cap \mathbb{S} = \emptyset$. For both cases, the rate of change $\frac{\partial x_i}{\partial \alpha}$ is defined by equation \eqref{eq_gig_definition}.

For $\{i, j\} \subset \mathbb{S}$,
\begin{equation}\label{eq_der_eq}
\begin{aligned}
\frac{\partial x_i}{\partial \alpha} &= x_i^I - x_i^B &&\qquad \text{ by \eqref{eq_gig_definition}} \\
& = x_j^I - x_j^B &&\qquad  \text{ by \eqref{eq_sym_val_at_start}} \\
& = \frac{\partial x_j}{\partial \alpha} &&\qquad  \text{ by \eqref{eq_gig_definition}}.
\end{aligned}
\end{equation}

For $\{i, j\} \cap \mathbb{S} = \emptyset$,
\begin{equation}\label{eq_der_0}
\frac{\partial x_i}{\partial \alpha} = \frac{\partial x_j}{\partial \alpha} = 0 \qquad  \text{ by \eqref{eq_gig_definition}}.
\end{equation}

By \eqref{eq_der_eq} and \eqref{eq_der_0}, it can be concluded that the rates of change of $x_i$ and $x_j$ are equal if $x_i = x_j$, Q.E.D. for req. 2.

Equations \eqref{eq_req_1}, \eqref{eq_der_eq} and \eqref{eq_der_0} prove that, in the context of symmetry preservation, the symmetric variables are equal at every point of the integration path. According to \emph{Lemma 2}, such a path is symmetry preserving. Q.E.D.

%% file: tex/pseudocode.tex
\algnewcommand{\Inputs}[1]{%
  \State \textbf{Inputs:}
  \Statex \hspace*{\algorithmicindent}\parbox[t]{.8\linewidth}{\raggedright #1}
}

\algnewcommand{\Initialize}[1]{%
  \State \textbf{Initialize:}
  \Statex \hspace*{\algorithmicindent}\parbox[t]{.8\linewidth}{\raggedright #1}
}

\section{Guided IG Pseudocode}

\begin{algorithm}[!thb]
\caption{Computation of Unbounded Guided IG\label{alg:gig}}
\begin{algorithmic}
\Inputs{
Input example $X^I$, \\
Baseline $X^B$, \\
Number of steps $T>0$, \\
Gradient of the function $grad(x)$,  \\
Target fraction of features to change at each step $p \in (0, 1]$.
}
\\
\Initialize{
\State $d_{total} \gets ||X^B-X^I||_1$
\State $x \gets X^B$
\State $attr \gets zeros(\text{size of $X^I$})$
}
\\
\For{$t \gets 1$ to $T$}
\State $y \gets grad(x)$
\Repeat
\State $y_i \gets \infty, \quad \forall i \: | \: x_i = X^I_i$ \Comment{as in \eqref{eq_y}}
\State $d_{target} \gets$ $d_{total} (1-\frac{t}{T})$
\State $d_{current} \gets$ $||x - X^I||_1$
\If {$d_{target}$ = $d_{current}$}
\State \textbf{break}
\EndIf
\\
\Comment{Assign to $S$ the $p$ fraction of features with the lowest absolute gradient values:}
\State $S \gets \{i \: | \: |y_i| \leq fraction(p, |y|)\}$  
\State $d_{S} \gets$ $\sum_{i \in S}|x_i - X_i^I|$
\State $\delta \gets \frac{d_{current}-d_{target}}{d_S}$
\State temp $\gets x$
\If {$\delta > 1$}
\State $x_i \gets X_i^I, \quad \forall i \in S$
\Else
\State $x_i \gets (1-\delta) x_i + \delta X_i^I, \quad \forall i \in S$
\EndIf
\State $y_i = 0, \quad \forall i = \infty$ \Comment{avoid mul. by $\infty$.}
\State $attr_i \gets attr_i + (x_i-temp_i) y_i, \quad \forall i \in S$ 
\Until{$\delta \leq 1$}
\EndFor
\State \Return $attr$
\end{algorithmic}
\end{algorithm}

%% file: tex/galleries.tex
\section{Example Results} \label{sec:gallery}
\begin{figure*}
  \centering
  \includegraphics[width=1\linewidth]{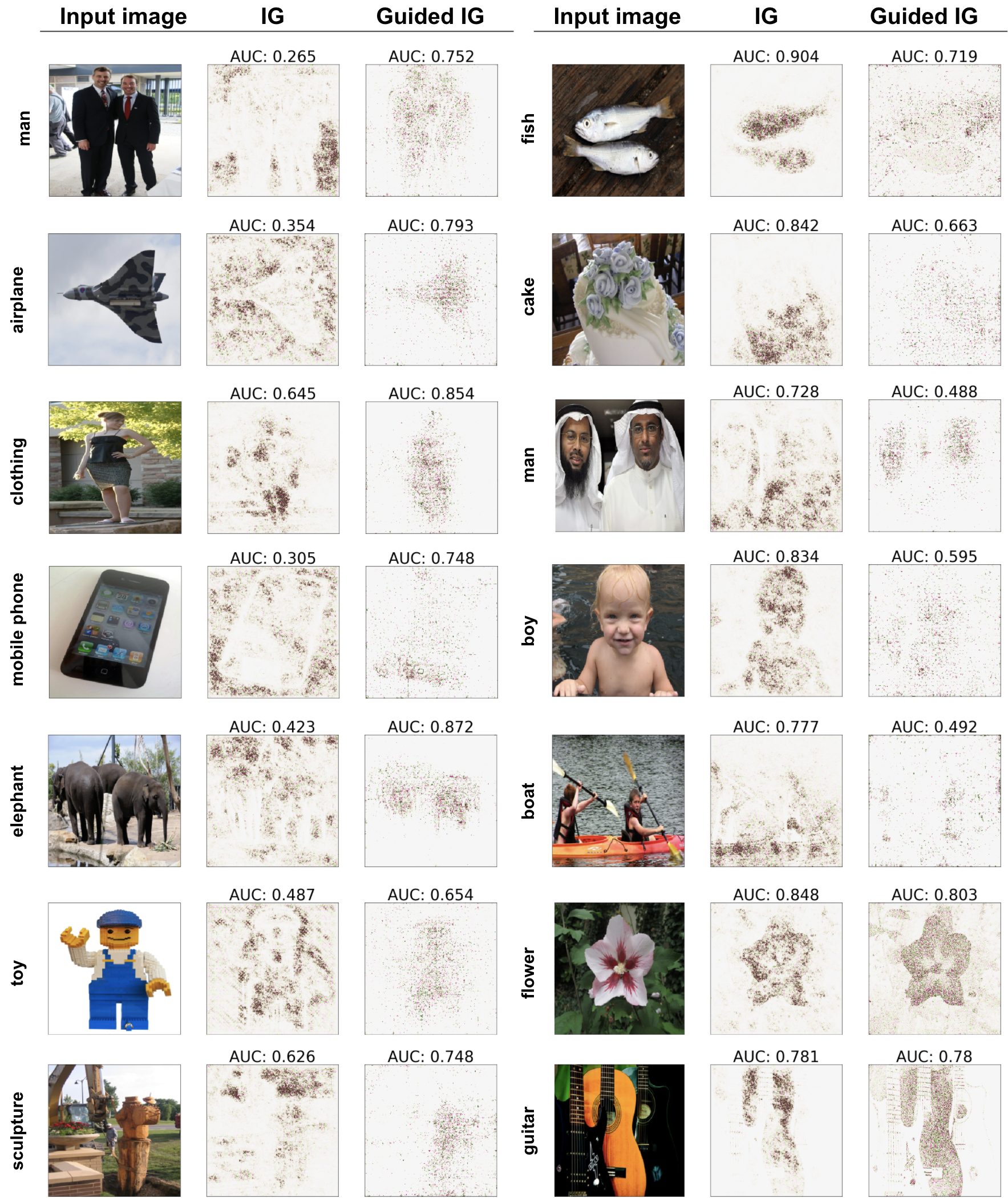}
  \caption{Comparing feature attribution for Integrated Gradients and Guided IG.  All images are from the OpenImages V5 dataset, and a black baseline was used for both methods. The left set of images were chosen randomly from images where the AUC for Guided IG was above average (AUC $>$ 0.630), and where the AUC for Guided IG was greater than IG. The right set of images were chosen randomly from images where the AUC for IG was above average (AUC $>$ 0.557), and images where the AUC for IG was greater than Guided IG.}
 \label{fig:gallery_all}
\end{figure*}

\paragraph{Comparing IG and GuidedIG} Fig.~\ref{fig:gallery_all} shows a selection of examples results for IG, and Guided IG.

\begin{figure}[!hbt]
  \centering
  \begin{subfigure}[b]{.95\linewidth}
  \includegraphics[width=\linewidth]{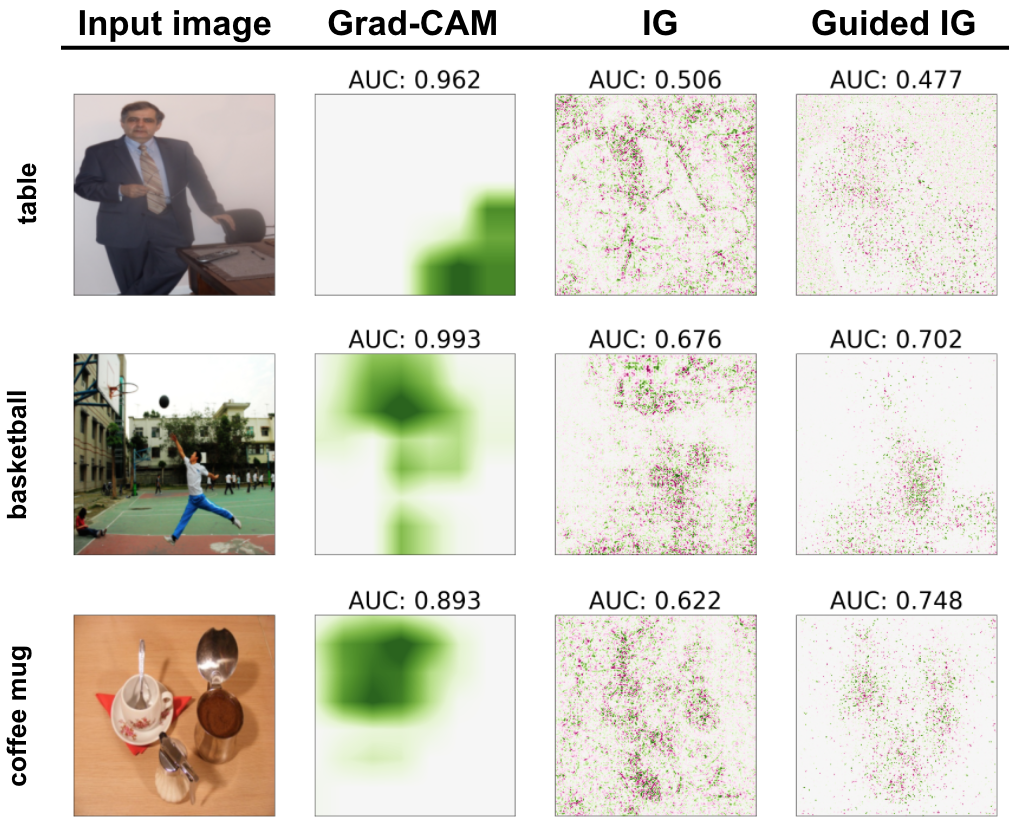}
  \end{subfigure}
  \caption{This shows images where Grad-CAM performs better in AUC localization metrics compared to IG and GuidedIG.  Images and segmentations are from the OpenImages V5 Dataset, and the model used was an InceptionV2 model trained on the ImageNet dataset. For the gradient-based methods, attributions were calculated separately with a black and white baseline and were averaged.}
  \label{fig:gradcam-win}
\end{figure}

\begin{figure}[!hbt]
  \centering
  \begin{subfigure}[b]{.95\linewidth}
  \includegraphics[width=\linewidth]{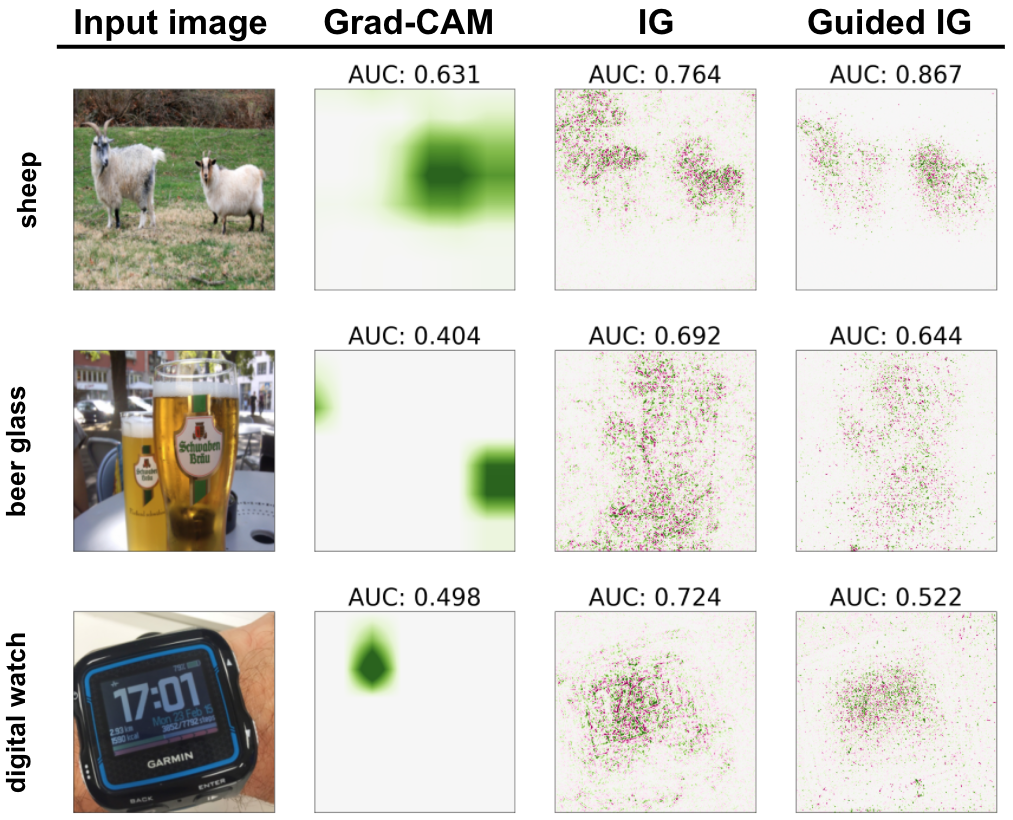}
  \end{subfigure}
  \caption{This shows images where IG performs better in AUC localization metrics compared to GradCAM. Images and segmentations are from the OpenImages V5 Dataset, and the model used was an InceptionV2 model trained on the ImageNet dataset. For the gradient-based methods, attributions were calculated separately with a black and white baseline and were averaged.}
  \label{fig:gradient-win}
\end{figure}

\paragraph{Comparisons with GradCAM} Fig.~\ref{fig:gradcam-win} shows examples where GradCAM outperforms IG and GuidedIG (based on AUC scores). Fig.~\ref{fig:gradient-win} shows examples where IG outperforms GradCAM.

%% file: guided_ig.bbl
\begin{thebibliography}{10}\itemsep=-1pt

\bibitem{AumannShapley}
R.~J. AUMANN and L.~S. SHAPLEY.
\newblock {\em Values of Non-Atomic Games}.
\newblock Princeton University Press, 1974.

\bibitem{Bach2015OnPE}
S. Bach, Alexander Binder, Gr{\'e}goire Montavon, F. Klauschen, K. M{\"u}ller,
  and W. Samek.
\newblock On pixel-wise explanations for non-linear classifier decisions by
  layer-wise relevance propagation.
\newblock {\em PLoS ONE}, 10, 2015.

\bibitem{bau2017network}
David Bau, Bolei Zhou, Aditya Khosla, Aude Oliva, and Antonio Torralba.
\newblock Network dissection: Quantifying interpretability of deep visual
  representations.
\newblock In {\em Proceedings of the IEEE conference on computer vision and
  pattern recognition}, pages 6541--6549, 2017.

\bibitem{BMBMS16}
Alexander Binder, Gr{\'e}goire Montavon, Sebastian Bach, Klaus-Robert
  M{\"u}ller, and Wojciech Samek.
\newblock Layer-wise relevance propagation for neural networks with local
  renormalization layers.
\newblock {\em CoRR}, 2016.

\bibitem{BylinskiiJOTD16}
Zoya Bylinskii, Tilke Judd, Aude Oliva, Antonio Torralba, and Fr{\'{e}}do
  Durand.
\newblock What do different evaluation metrics tell us about saliency models?
\newblock {\em CoRR}, abs/1604.03605, 2016.

\bibitem{eyepacs_2009}
Jorge Cuadros and George Bresnick.
\newblock Eyepacs: an adaptable telemedicine system for diabetic retinopathy
  screening.
\newblock {\em Journal of Diabetes Science and Technology}, 3(3):509--516,
  2009.

\bibitem{Dombrowski2019ExplanationsCB}
Ann-Kathrin Dombrowski, M. Alber, Christopher~J. Anders, M. Ackermann, K.
  M{\"u}ller, and P. Kessel.
\newblock Explanations can be manipulated and geometry is to blame.
\newblock In {\em NeurIPS}, 2019.

\bibitem{fong2019understanding}
Ruth Fong, Mandela Patrick, and Andrea Vedaldi.
\newblock Understanding deep networks via extremal perturbations and smooth
  masks.
\newblock In {\em Proceedings of the IEEE International Conference on Computer
  Vision}, pages 2950--2958, 2019.

\bibitem{fong2017interpretable}
Ruth~C Fong and Andrea Vedaldi.
\newblock Interpretable explanations of black boxes by meaningful perturbation.
\newblock In {\em Proceedings of the IEEE International Conference on Computer
  Vision}, pages 3429--3437, 2017.

\bibitem{Friedman2004PathsAC}
E. Friedman.
\newblock Paths and consistency in additive cost sharing.
\newblock {\em International Journal of Game Theory}, 32:501--518, 2004.

\bibitem{Ghorbani2017}
Amirata Ghorbani, Abubakar Abid, and James Zou.
\newblock Interpretation of neural networks is fragile.
\newblock {\em Proceedings of the AAAI Conference on Artificial Intelligence},
  33, 10 2017.

\bibitem{kapishnikov2019xrai}
Andrei Kapishnikov, Tolga Bolukbasi, Fernanda Vi{\'e}gas, and Michael Terry.
\newblock Xrai: Better attributions through regions.
\newblock In {\em Proceedings of the IEEE International Conference on Computer
  Vision}, pages 4948--4957, 2019.

\bibitem{kim2017interpretability}
Been Kim, Martin Wattenberg, Justin Gilmer, Carrie Cai, James Wexler, Fernanda
  Viegas, and Rory Sayres.
\newblock Interpretability beyond feature attribution: Quantitative testing
  with concept activation vectors (tcav), 2017.

\bibitem{krause_2018}
J. Krause, V. Gulshan, E. Rahimy, P. Karth, K. Widner, G.~S. Corrado, L. Peng,
  and D.~R. Webster.
\newblock {{G}rader {V}ariability and the {I}mportance of {R}eference
  {S}tandards for {E}valuating {M}achine {L}earning {M}odels for {D}iabetic
  {R}etinopathy}.
\newblock {\em Ophthalmology}, 125(8):1264--1272, 08 2018.

\bibitem{Kuznetsova_2020}
Alina Kuznetsova, Hassan Rom, Neil Alldrin, Jasper Uijlings, Ivan Krasin, Jordi
  Pont-Tuset, Shahab Kamali, Stefan Popov, Matteo Malloci, Alexander
  Kolesnikov, and et al.
\newblock The open images dataset v4.
\newblock {\em International Journal of Computer Vision}, 128(7):1956–1981,
  Mar 2020.

\bibitem{mahendran2015understanding}
Aravindh Mahendran and Andrea Vedaldi.
\newblock Understanding deep image representations by inverting them.
\newblock In {\em Proceedings of the IEEE conference on computer vision and
  pattern recognition}, pages 5188--5196, 2015.

\bibitem{miglani2020investigating}
Vivek Miglani, Narine Kokhlikyan, Bilal Alsallakh, Miguel Martin, and Orion
  Reblitz-Richardson.
\newblock Investigating saturation effects in integrated gradients, 2020.

\bibitem{Mordvintsev2015InceptionismGD}
A. Mordvintsev, Christopher Olah, and M. Tyka.
\newblock Inceptionism: Going deeper into neural networks.
\newblock 2015.

\bibitem{petsiuk2018rise}
Vitali Petsiuk, Abir Das, and Kate Saenko.
\newblock Rise: Randomized input sampling for explanation of black-box models,
  2018.

\bibitem{ribeiro2016should}
Marco~Tulio Ribeiro, Sameer Singh, and Carlos Guestrin.
\newblock Why should i trust you?: Explaining the predictions of any
  classifier.
\newblock In {\em Proceedings of the 22nd ACM SIGKDD international conference
  on knowledge discovery and data mining}, pages 1135--1144. ACM, 2016.

\bibitem{russakovsky_imagenet_2014}
Olga Russakovsky, Jia Deng, Hao Su, Jonathan Krause, Sanjeev Satheesh, Sean Ma,
  Zhiheng Huang, Andrej Karpathy, Aditya Khosla, Michael Bernstein,
  Alexander~C. Berg, and Li Fei-Fei.
\newblock {ImageNet} {Large} {Scale} {Visual} {Recognition} {Challenge}.
\newblock {\em arXiv:1409.0575 [cs]}, Sept. 2014.
\newblock arXiv: 1409.0575.

\bibitem{selvaraju2017grad}
Ramprasaath~R Selvaraju, Michael Cogswell, Abhishek Das, Ramakrishna Vedantam,
  Devi Parikh, and Dhruv Batra.
\newblock Grad-cam: Visual explanations from deep networks via gradient-based
  localization.
\newblock In {\em Proceedings of the IEEE International Conference on Computer
  Vision}, pages 618--626, 2017.

\bibitem{shapley1953value}
Lloyd~S Shapley.
\newblock A value for n-person games.
\newblock {\em Contributions to the Theory of Games}, 2(28):307--317, 1953.

\bibitem{SVZ13}
Karen Simonyan, Andrea Vedaldi, and Andrew Zisserman.
\newblock Deep inside convolutional networks: Visualising image classification
  models and saliency maps.
\newblock {\em CoRR}, 2013.

\bibitem{Smilkov2017SmoothGradRN}
Daniel Smilkov, Nikhil Thorat, Been Kim, Fernanda~B. Vi{\'e}gas, and Martin
  Wattenberg.
\newblock Smoothgrad: removing noise by adding noise.
\newblock {\em ArXiv}, abs/1706.03825, 2017.

\bibitem{springenberg2014striving}
Jost~Tobias Springenberg, Alexey Dosovitskiy, Thomas Brox, and Martin
  Riedmiller.
\newblock Striving for simplicity: The all convolutional net, 2014.

\bibitem{sturmfels2020visualizing}
Pascal Sturmfels, Scott Lundberg, and Su-In Lee.
\newblock Visualizing the impact of feature attribution baselines.
\newblock {\em Distill}, 2020.
\newblock https://distill.pub/2020/attribution-baselines.

\bibitem{sundararajan2020many}
Mukund Sundararajan and Amir Najmi.
\newblock The many shapley values for model explanation.
\newblock In {\em International Conference on Machine Learning}, pages
  9269--9278. PMLR, 2020.

\bibitem{Sundararajan2017AxiomaticAF}
Mukund Sundararajan, Ankur Taly, and Qiqi Yan.
\newblock Axiomatic attribution for deep networks.
\newblock In {\em ICML}, 2017.

\bibitem{Xu_2020_CVPR}
Shawn Xu, Subhashini Venugopalan, and Mukund Sundararajan.
\newblock Attribution in scale and space.
\newblock In {\em Proceedings of the IEEE/CVF Conference on Computer Vision and
  Pattern Recognition (CVPR)}, June 2020.

\bibitem{Zeiler_2014}
Matthew~D. Zeiler and Rob Fergus.
\newblock Visualizing and understanding convolutional networks.
\newblock {\em Lecture Notes in Computer Science}, page 818–833, 2014.

\end{thebibliography}
